\documentclass{nle}
\usepackage{booktabs}
\usepackage{amsmath}
\usepackage{amssymb}
\usepackage{url}
\usepackage{floatrow}
\usepackage{algorithm}
\usepackage[noend]{algorithmic}
\usepackage{graphicx}
\usepackage{multirow}
\usepackage{verbatim}
\usepackage{rotating}
\usepackage{todonotes}
\usepackage{enumitem}
\usepackage[font=small,labelfont=bf]{caption}
\newcommand{\parm}{\mathord{\bullet}}

\makeatletter
\newcommand\@ptsize{1}
\makeatother
\usepackage{setspace}

\newcommand{\argmax}{\operatornamewithlimits{argmax}\limits}

\newfloatcommand{capbtabbox}{table}[][\FBwidth]

\title[Distributional Semantics for Enriching Lexical Semantic Resources]
      {A Framework for Enriching Lexical Semantic Resources with Distributional Semantics}
\author[Biemann et al.]
       {%
       	Chris Biemann$^1$, Stefano Faralli$^2$, 
        Alexander Panchenko$^1$, Simone P.\ Ponzetto$^2$\\
        $^1$Language Technology Group, Department of Informatics, Faculty of Mathematics,\\ Informatics, and Natural Sciences, Universit{\"a}t Hamburg, Germany\\ \{biemann,panchenko\}@informatik.uni-hamburg.de\\ 
        $^2$Data and Web Science Group, School of Business Informatics and Mathematics, \\ Universit{\"a}t Mannheim, Germany\\ \{stefano,simone\}@informatik.uni-mannheim.de}

\newcommand{\term}[1]{\textsf{#1}}              
\newcommand{\sense}[1]{\textsf{#1}}            

\begin{document}

\label{firstpage}
\maketitle

\begin{abstract}
We present an approach to combining distributional semantic representations induced from text corpora with manually constructed lexical-semantic networks. While both kinds of semantic resources are available with high lexical coverage, our aligned resource combines the domain specificity and availability of contextual information from distributional models with the conciseness and high quality of manually crafted lexical networks.  
We start with a distributional representation of induced senses of vocabulary terms, which are accompanied with rich context information given by related lexical items. We then automatically disambiguate such representations to obtain a full-fledged proto-conceptualization, i.e.\ a typed graph of induced word senses. In a final step, this proto-conceptualization is aligned to a lexical ontology, resulting in a hybrid aligned resource. Moreover, unmapped induced senses are associated with a semantic type in order to connect them to the core resource.
Manual evaluations against ground-truth judgments for different stages of our method as well as an extrinsic evaluation on a knowledge-based Word Sense Disambiguation benchmark all indicate the high quality of the new hybrid resource. Additionally, we show the benefits of enriching top-down lexical knowledge resources with bottom-up distributional information from text for addressing high-end knowledge acquisition tasks such as cleaning hypernym graphs and learning taxonomies from scratch.
\end{abstract}

\section{Introduction}
\label{sec:intro}

Automatic enrichment of semantic resources is an important problem \cite{Biemann05,JurgensP16} as it attempts to alleviate the extremely costly process of \textit{manual} lexical resource construction.  Distributional semantics~\cite{turney10,baroni14,clark15} provides an alternative \textit{automatic} meaning representation framework that has been shown to benefit many text-understanding applications. 

Recent years have witnessed an impressive amount of work on combining the symbolic semantic information available in manually constructed lexical resources with distributional information, where words are usually represented as dense numerical vectors, a.k.a. embeddings. Examples of such approaches include methods that modify the Skip-gram model \cite{mikolov13} to learn sense embeddings \cite{chen14} based on the sense inventory of WordNet, methods that learn embeddings of synsets as given in a lexical resource \cite{rothe15} or methods that acquire  word vectors  by applying random walks over lexical resources to learn a neural language model \cite{goikoetxeaSA15}. Besides, alternative approaches like NASARI \cite{camachocollados15} and MUFFIN \cite{camachocollados15b} looked at ways to produce joint lexical and semantic vectors for a common representation of word senses in text and in lexical resources. Retrofitting approaches, e.g. ~\cite{faruqui15b}, efficiently ``consume'' lexical resources as input in order to improve the quality of word embeddings, but do not add anything to the resource itself. Other approaches, such as AutoExtend~\cite{rothe15}, NASARI and MUFFIN, learn vector representations for existing synsets that can be added to the resource, however are not able to add missing senses discovered from text.  

In these contributions, the benefits of such hybrid knowledge sources for tasks in computational semantics such as semantic similarity and Word Sense Disambiguation (WSD) \cite{Navigli2009} have been extensively demonstrated. However, none of the existing approaches, to date, have been designed to use distributional information for the enrichment of lexical semantic resources, i.e. adding new symbolic entries. 

In this article, we consequently set out to filling this gap by developing \textbf{a framework for enriching lexical semantic resources with distributional semantic models}. The goal of such framework is the creation of a resource that brings together the `best of both worlds', namely  the \emph{precision} and \emph{interpretability} from the lexicographic tradition that describes senses and semantic relations manually, and the \emph{versatility} of data-driven, corpus-based approaches that derive senses automatically. \\

A lexical resource enriched with additional knowledge induced from text can boost the performance of natural language understanding tasks like WSD or Entity Linking \cite{mihalcea07,Rospocher2016}, where it is crucial to have a comprehensive list of word senses as well as the means to assign the correct of many possible senses for a given word in context. 

Consider, for instance, the sentence:
\begin{flushleft}
``Regulator of calmodulin signalling (RCS) knockout \term{mice} display anxiety-like behavior and motivational deficits''.\footnote{\url{http://www.ncbi.nlm.nih.gov/pmc/articles/PMC3622044}} \hfill
\end{flushleft}
No synset for ``RCS'' can be found in either WordNet \cite{Fellbaum:98} or BabelNet \cite{navigli12}, yet it is distributionally related to other bio-medical concepts and thus can help to disambiguate the ambiguous term \term{mice} to its `animal' meaning, as opposed to the `computer peripheral device'. 

Our approach yields a hybrid resource that combines symbolic and statistical meaning representations while i) staying purely on the lexical-symbolic level, ii) explicitly distinguishing word senses, and iii) being human readable. These properties are crucial to be able to embed such a resource in an explicit semantic data space such as, for instance, the Linguistic Linked Open Data ecosystem \cite{Chiarcos12}. According to \cite{norvig16}, the Semantic Web and natural language understanding are placed at the heart of current efforts to understand the Web on a large scale.

We take the current line of research on hybrid semantic representations one step forward by presenting a methodology for inducing distributionally-based sense representations from text, and for linking them to a reference lexical resource. Central to our method is a novel sense-based distributional representation that we call proto-conceptualization (PCZ). A PCZ is a repository of word senses induced from text, where each sense is represented with related senses, hypernymous senses, and aggregated clues for contexts in which the respective sense occurs in text. Besides, to further improve interpretability, each sense has an associated image. 
We build a bridge between the PCZ and lexical semantic networks by establishing a mapping between these two kinds of resources.\footnote{We use WordNet and BabelNet, however our approach can be used with similar resources, e.g.\ those listed at \url{http://globalwordnet.org/wordnets-in-the-world}. } This results in a new knowledge resource that we refer to as \emph{hybrid aligned resource}: here, senses induced from text are aligned to a set of synsets from a reference lexical resource, whereas induced senses that cannot be matched are included as additional synsets. By linking our distributional representations to a repository of symbolically-encoded knowledge, we are able to complement wide-coverage statistical meaning representations with explicit relational knowledge as well as to extend the coverage of the reference lexical resource based on the senses induced from a text collection. The main contributions of this article are listed as follows:

\begin{itemize}[leftmargin=4mm]

\item \textbf{A framework for enriching lexical semantic resources}: we present a methodology for combining information from distributional semantic models with manually constructed lexical semantic resources.

\item \textbf{A hybrid lexical semantic resource}: we apply our framework to produce a novel hybrid resource obtained by linking disambiguated distributional lexical semantic networks to WordNet and BabelNet. 

\item \textbf{Applications of the hybrid resource}: besides \emph{intrinsic} evaluations of our approach, we test the utility of our resource \emph{extrinsically} on the tasks of word sense disambiguation and taxonomy induction, demonstrating the benefits of combining distributional and symbolic lexical semantic knowledge.

\end{itemize}

\medskip
\noindent
The remainder of this article is organized as follows: we first review related work in Section \ref{sec:related} and provide an overview of our framework to build a hybrid aligned resource from distributional semantic vectors and a reference knowledge graph in Section \ref{sec:workflow}. Next, we provide details on our methodology to construct PCZs and to link them to a lexical resource in Sections \ref{sec:DDTdesc} and \ref{sec:linking}, respectively. In Section \ref{sec:experiments}, we benchmark the quality of our resource in different evaluation settings, including an intrinsic and an extrinsic evaluation on the task of knowledge-based WSD using a dataset from a SemEval task. Section \ref{sec:applications} provides two application-based evaluations that demonstrate how our hybrid resource can be used for taxonomy induction. We conclude with final remarks and future directions in Section \ref{sec:conclusions}.

\section{Related Work}
\label{sec:related}

\subsection{Automatic Construction of Lexical Semantic Resources}
In the past decade, large efforts have been undertaken to research the automatic acquisition of machine-readable knowledge on a large scale by mining large repositories of textual data \cite{banko07,carlson10,fader11,faruqui15a}. At this, collaboratively constructed resources have been exploited, used either in isolation \cite{bizer09,ponzetto11,nastase12}, or complemented with manually assembled knowledge sources \cite{Suchaneketal:08,navigli12,gurevych12,hoffart13}. As a result of this, recent years have seen a remarkable renaissance of knowledge-rich approaches for many different artificial intelligence tasks \cite{hovy13a}. Knowledge contained within these very large knowledge repositories, however, has major limitations in that these resources typically do not contain any linguistically grounded probabilistic representation of concepts, instances, and their attributes -- namely, the bridge between wide-coverage conceptual knowledge and its instantiation within natural language texts. While there are large-scale lexical resources derived from large corpora such as ProBase \cite{Probase12}, these are usually not sense-aware but conflate the notions of term and concept. With this work, we provide a framework that aims at augmenting any of these wide-coverage knowledge sources with distributional semantic information, thus extending them with text-based contextual information. 

Another line of research has looked at the problem of Knowledge Base Completion \cite{nickel16} (KBC). Many approaches to KBC focus on exploiting other KBs \cite{wang2012cross,bryl2014learning} for acquiring additional knowledge, or rely on text corpora -- either based on distant supervision \cite{snow06,mintz2009distant,aprosio2013extending} or by rephrasing KB relations as queries to a search engine \cite{west2014knowledge} that returns results from the web as corpus. Alternative methods primarily rely on existing information in the KB itself \cite{bordes2011learning,jenatton2012latent,socher2013reasoning,klein17} to simultaneously learn continuous representations of KB concepts and KB relations by exploiting the KB structure as the ground truth for supervision, inferring additional relations from existing ones. Lexical semantic resources and text are  synergistic sources, as shown by complementary work from \cite{faruqui15b}, who improve the quality of semantic vectors based on lexicon-derived relational information. 

Here, we follow this intuition of combining structured knowledge resources with distributional semantic information from text, but focus instead on providing hybrid semantic representations for KB concepts and entities, as opposed to the classification task of KBC that aims at predicting additional semantic relations between known entities.

\subsection{Combination of Distributional Semantics with Lexical Resources}
Several prior approaches combined distributional information extracted from text with information available in lexical resources like e.g. WordNet. This includes a model \cite{yu-dredze:2014:P14-2} to learn word embeddings based on lexical relations of words from WordNet and PPDB~\cite{ganitkevitch-vandurme-callisonburch:2013:NAACL-HLT}. The objective function of this model combines that of the skip-gram model~\cite{mikolov13} with a term that takes into account lexical relations of target words. In work aimed at retrofitting word vectors \cite{faruqui15b}, a related approach was proposed that performs post-processing of word embeddings on the basis of lexical relations from lexical resources. Finally, \cite{pham-lazaridou-baroni:2015:ACL-IJCNLP} also aim at improving word vector representations by using lexical relations from WordNet, targeting similar representations of synonyms and dissimilar representations of antonyms. While all these three approaches show excellent performance on word relatedness evaluations, they do not model word senses -- in contrast to other work aimed instead at learning sense embeddings using the word sense inventory of WordNet \cite{jauhar-dyer-hovy:2015:NAACL-HLT}.

A parallel line of research has recently focused on learning unified statistical and symbolic semantic representations. Approaches aimed at providing unified semantic representations from distributional information and lexical resources have accordingly received an increasing level of attention \cite{chen14,rothe15,goikoetxeaSA15,camachocollados15,nietopina2016}, \emph{inter alia} (cf.\ also our introductory discussion in Section \ref{sec:intro}), and hybrid meaning representations have been shown to benefit challenging semantic tasks such as WSD and semantic similarity at word level and text level. 

All these diverse contributions indicate the benefits of hybrid knowledge sources for learning word and sense representations: here, we further elaborate along this line of research and develop a new hybrid resource that combines information from the knowledge graph with distributional sense representations that are human readable and easy to interpret, in contrast to dense vector representations, a.k.a. word embeddings like GloVe \cite{pennington14} or word2vec \cite{mikolov13}. As a result of this, we are able to provide, to the best of our knowledge, the first hybrid knowledge resource that is fully integrated and embedded within the Semantic Web ecosystem provided by the Linguistic Linked Open Data cloud \cite{Chiarcos12}. Note that this complementary to recent efforts aimed at linking natural language expressions in text with semantic relations found within LOD knowledge graphs \cite{krause15}, in that we focus instead on combining explicit semantic information with statistical, distributional semantic representations of concepts and entities into an augmented resource.

\section{Enriching Lexical Semantic Resources with Distributional Semantics}
\label{sec:workflow}

\begin{figure}[t!]
 \centering
 \includegraphics[width=1.0\columnwidth]{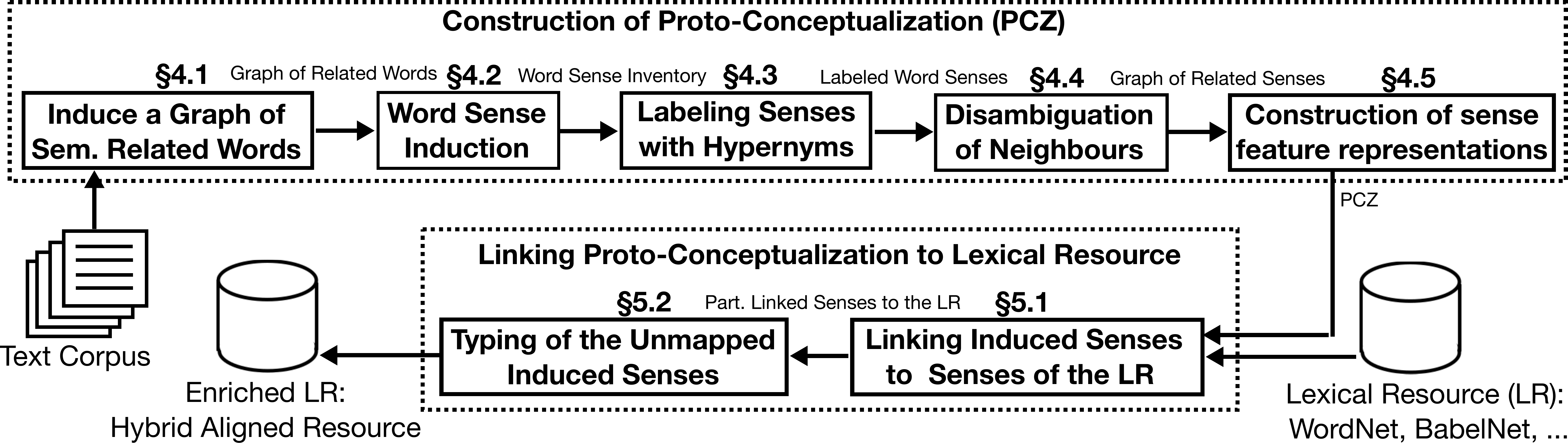}
 \caption{Overview of the proposed framework for enriching lexical resources: a distributional semantic model is used to construct a disambiguated distributional lexical semantic network (a proto-conceptualization, PCZ), which is subsequently linked to the lexical resource.}
 \label{fig:workflow}
\end{figure}

The construction of our hybrid aligned resource (HAR) builds upon methods used to link various manually constructed lexical resources to construct BabelNet~\cite{ponzetto2010knowledge} and UBY~\cite{gurevych12}, among others. In our method, however, linking is performed between two networks that are structurally similar, but have been constructed in two completely different ways: one resource is built using an unsupervised bottom-up approach from text corpora, while the second is constructed in a top-down manner using manual labor, e.g., codified knowledge from human experts such as lexicographers (WordNet). In particular, the method consists of two major phases, as illustrated in Figure \ref{fig:workflow}:

\begin{table}[t]
\footnotesize
\begin{tabular}{p{1.5cm}p{3.2cm}p{3cm}p{3cm}}
	\toprule
    \bf Word Sense & \bf Related Senses  & \bf Hypernyms & \bf Context Clues\\
    \midrule
    mouse\textbf{:0} &  rat\textbf{:0}, rodent\textbf{:0}, monkey\textbf{:0}, ... & animal\textbf{:0}, species\textbf{:1}, ... &  rat:conj\_and, white-footed:amod, ...\\
	\midrule
    mouse\textbf{:1} & keyboard\textbf{:1}, printer\textbf{:0}, computer\textbf{:0} ... & device\textbf{:1}, equipment\textbf{:3},  ... & click:-prep\_of, click:-nn, ....  \\
	\midrule
    keyboard\textbf{:0} & piano\textbf{:1}, synthesizer\textbf{:2}, organ\textbf{:0} ... & instrument\textbf{:2}, device\textbf{:3}, ... & play:-dobj, electric:amod, .. \\
	\midrule
    keyboard\textbf{:1} & keypad\textbf{:0}, mouse\textbf{:1}, screen\textbf{:1} ... & device\textbf{:1}, technology\textbf{:0} ... & computer:nn, qwerty:amod ... \\
    \bottomrule
\end{tabular}
\caption{Examples of entries of the PCZ resource for words \term{mouse} and \term{keyboard} after disambiguation of their related terms and hypernyms (Section \ref{sec:disambiguation}). Context clues are represented as typed dependency relations to context words in the input corpus, e.g. \term{keyboard:0} appears as direct object of "to play". Trailing numbers indicate automatically induced sense identifiers.}
\label{tab:DDTex}
\end{table}

\begin{enumerate}[leftmargin=4mm]

\item  \textbf{Construction of a proto-conceptualization (PCZ) from text}. Initially, a symbolic disambiguated distributional lexical semantic network, called a \emph{proto-conceptualization} (PCZ), is induced from a text corpus. To this end, we first create a sense inventory from a large text collection using graph-based word sense induction as provided by the JoBimText project~\cite{BiemannRiedl2013}. The resulting structure contains induced word senses, their yet un-disambiguated related terms, as well as context clues per term. First, we obtain sense representations by aggregating context clues over sense clusters. Second, we disambiguate related terms and hypernyms to produce a fully disambiguated resource where all terms have a sense identifier. Hence, the PCZ is a repository of corpus-induced word senses, where each sense is associated with a list of related senses, hypernymous senses, as well as aggregated contextual clues (Table~\ref{tab:DDTex}). 
 
\item \textbf{Linking a proto-conceptualization (PCZ) to a lexical resource (LR).} Next, we align the PCZ with an existing lexical resource (LR). In our work, we make use of lexical semantic resources such as WordNet and BabelNet featuring large vocabularies of lexical units with explicit meaning representations as well as semantic relations between these. In this phase we create a mapping between the two sense inventories from the PCZ and the LR, and combine them into a new extended sense inventory, our {\em Hybrid Aligned Resource} (HAR). Finally, to obtain a complete unified resource, we link the `orphan' PCZ senses for which no corresponding sense could be found by inferring their type (i.e., their most specific generalization) in the LR.

\end{enumerate}
In the following sections, we present each stage of our approach in detail.

\section{Construction of a Proto-Conceptualization}
\label{sec:DDTdesc}

Our method for proto-conceptualization (PCZ) construction consists of the four steps illustrated in the upper half of Figure~\ref{fig:workflow}: inducing a graph of semantically related words (Section \ref{sec:dt}); word sense induction (Section \ref{sec:wsi}); labeling of clusters with hypernyms and images (Section \ref{sec:hypernyms}), and disambiguation of related words and hypernyms with respect to the induced sense inventory (Section \ref{sec:disambiguation}). Further, we describe an additional property of our PCZs, namely the availability of corpus-derived context clues (Section \ref{sec:sensevec}), as well as alternatives ways to construct PCZs on the basis of dense vector representations (Section \ref{sec:densevec}).

\begin{figure}
\begin{center}
\includegraphics[width=1.0\textwidth]{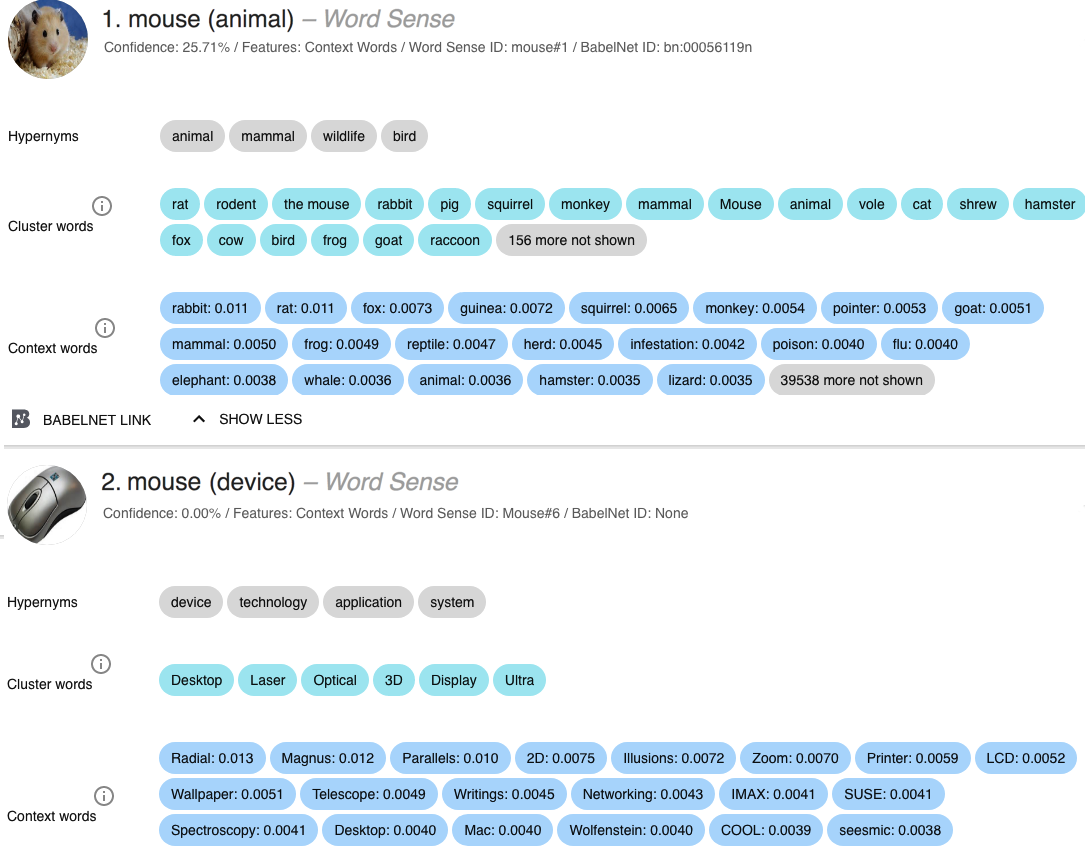}
\end{center}
\caption{Word sense representations of the word \term{mouse} induced from text generated using the online demo at \texttt{http://jobimtext.org/wsd}. The sense labels (\term{device} and \term{animal}) are obtained automatically based on cluster labeling with hypernyms. The images associated with the senses are retrieved with a search engine using the queries: \term{mouse device} and \term{mouse animal}. Note the ``BabelNet Link'' button, leading to the sense in BabelNet linked to the induced sense with the algorithm described in Section \ref{sec:linking}.
}
\label{fig:demo}
\end{figure}

The PCZ resource with a pipeline as outlined in Figure~\ref{fig:workflow} consists of word senses induced from a corpus. For each word sense, similar and superordinate terms are disambiguated with respect to the induced sense inventory: the structure of a PCZ resembles that of a lexical semantic resource. Sense distinctions and distributions depend on the training corpus, which causes the resource to adapt to its domain. 
In contrast to manually created resources, each sense also contains context clues that allow disambiguating polysemous terms in context. Table \ref{tab:DDTex} shows example senses for the terms \term{mouse} and \term{keyboard}. Note that PCZs may contain many entries for the same word, e.g. \term{mouse} has two senses, the `animal' and the `computer device', respectively. The context clues are not disambiguated, since they are meant for directly matching (undisambiguated) textual context. PCZs  can be interpreted by humans at two levels, as exemplified in Figure~\ref{fig:demo}.

\begin{enumerate}[leftmargin=4mm]
\item The \textbf{word sense inventory} is interpretable due to the use of the hypernyms, images and related senses.

\item The \textbf{sense feature representation} is interpretable due to the use of the sparse context clues relevant to the sense. 

\end{enumerate}
Note that while in our experiments we rely on a count-based sparse distributional model, the PCZ is a symbolic structure that can be also constructed using alternative distributional models, e.g. word and sense embeddings (cf.\ Section~\ref{sec:densevec}).

\subsection{Inducing a Graph of Semantically Related Words}
\label{sec:dt}

The goal of this first stage is to build a graph of semantically related words, with edges such as~{\small \textsf{(mouse, keyboard, 0.78)}}, i.e., a distributional thesaurus (DT)~\cite{Lin98}. To induce such graph in an unsupervised way we, rely on a count-based approach to distributional semantics based on the \textit{JoBimText} framework~\cite{BiemannRiedl2013}. Each word  is represented by a bag of syntactic dependencies such as \textsf{conj\_and(rat, $\parm$)} or  \textsf{prep\_of(click,$\parm$)}, extracted from the Stanford Dependencies~\cite{de2006generating} obtained with the PCFG model of the Stanford parser~\cite{klein-manning:2003:ACL}.  

Features of each word are weighted and ranked using the Local Mutual Information (LMI) metric~\cite{Evert2005}. Subsequently, these word  representations are pruned keeping 1,000 most salient features per word and 1,000 most salient words per feature. The pruning reduces computational complexity and noise \cite{riedl2016phd}. Finally, word similarities are computed as the number of common features for two words. This is, again, followed by a pruning step in which for every word, only the 200 most similar terms are kept. The resulting graph of word similarities is browsable online~\cite{ruppertEtAl15}.\footnote{Word and sense representations used in our experiments can be inspected by selecing the ``Stanford (English)'' model in the JoBimViz demo at \url{http://jobimtext.org/jobimviz/}.}

\begin{table}
\footnotesize  
\centering

\begin{tabular}{p{9.5cm}rr} 
\toprule 
\bf Method & \bf high & \bf low \\ \midrule

Lin's similarity \cite{Lin98} & 0.2872 & 0.2291 \\
$t$-test \cite{curran2002ensemble} & 0.2589 & 0.2067 \\
Skip-gram \cite{mikolov13} & 0.2548 & 0.2068 \\
Skip-gram with dependency features \cite{levy-goldberg:2014:P14-2} & 0.2632 & 0.1992  \\
LMI with trigram features \cite{Riedl13scale} & 0.2621 & 0.2003 \\
LMI with dependency features \cite{Riedl13scale} & \bf 0.2933 & \bf 0.2337 \\
\bottomrule
\end{tabular}
\caption{Comparison of state-of-the-art count- and prediction-based methods to distributional semantics on the basis of the average of the averaged similarity scores between each term in the DT and its top-10 most similar terms using the WordNet path similarity measure (higher means better) averaged over 1000 high- and low-frequency words. In this article, we use ``LMI with dependency features'' as the similarity function.}
\label{tab:word2vecf}
\end{table}

There are many possible ways to compute a graph of semantically similar words, including count-based approaches, such as \cite{Lin98,curran2002ensemble} or prediction-based approaches, such as word2vec~\cite{mikolov13}, GloVe~\cite{pennington14}, and word2vecf~\cite{levy-goldberg:2014:P14-2}. Here, we opt for a count-based approach to distributional semantics based on LMI based on two considerations, namely their higher quality of similarity scores and their interpretability. 

A thorough experimental comparison of different approaches to computing distributional semantic similarity to build a distributional thesaurus is presented by Riedl (2016, Section 5.7.4) using the WordNet taxonomy as a gold standard. In this evaluation, different DTs are compared by computing, for each term, the average similarity between the term itself and its $k$ most similar terms (based on the DT) using the WordNet path-based similarity measure \cite{pedersen04}. The overall similarity of the DT with the ground-truth taxonomy (e.g., WordNet) is then given by the average similarity score across all terms. Using this evaluation framework, Riedl is able to compare a wide range of different approaches for the construction of a weighted similarity graph, including state-of-the-art approaches based on sparse vector representations~\cite{Lin98,curran2002ensemble}, as well as dense representations based on word2vec~\cite{mikolov13} and word2vecf, which makes use of dependency-based features~\cite{levy-goldberg:2014:P14-2}. In his experiment, all methods were trained on the same corpus, and all dependency-based models, including the Skip-gram approach trained with the word2vecf tool~\cite{levy-goldberg:2014:P14-2}, used the same feature representations. 

We report some of the results from Riedl's experiments in Table~\ref{tab:word2vecf}. In this experiment, 1,000 infrequent and 1,000 frequent nouns proposed by~\cite{weeds2004characterising} were used. Dependency-based models (all models except the Skip-gram) used syntactic features extracted using the Stanford Parser. In addition to this dependency-based model, we report results of the same model based on trigram features, where a context is formed by the concatenation of the two adjacent words.  All models were trained on the 105 million sentences of newspaper data described in Section~\ref{sec:datasets}. Further details of the experiment, e.g. parameters of the models, are available in Riedl (2016, Section 5.7.4).

The performance figures indicate that the method we use here yields the overall best performance in terms of semantic similarity compared to other count-based or word-embedding approaches (including both word2vec and word2vecf). Besides, the results generally indicate the advantage of using dependency-based context representations over the bag-of-words representations. This is in line with prior studies on semantic similarity \cite{pado2007dependency,van2010mining,panchenko2012hybrid}. For this reason, we use syntactic features in our experiments but would like to stress that the overall framework also allows simpler context representations, giving rise to its application to resource-poor languages.

The second reason for using the LMI approach to compute a graph of semantically related words is the fact that the resulting word representations are human-interpretable, since words are represented by sparse features -- as opposed to dense features such as those found within word embeddings. Besides being a value on its own, this feature enables a straightforward implementation of word sense disambiguation methods on the basis of the learned representations~\cite{panchenko2017sense,panchenko2017unsup}.

\subsection{Word Sense Induction}
\label{sec:wsi}

\begin{figure}
\begin{center}
\includegraphics[width=\textwidth]{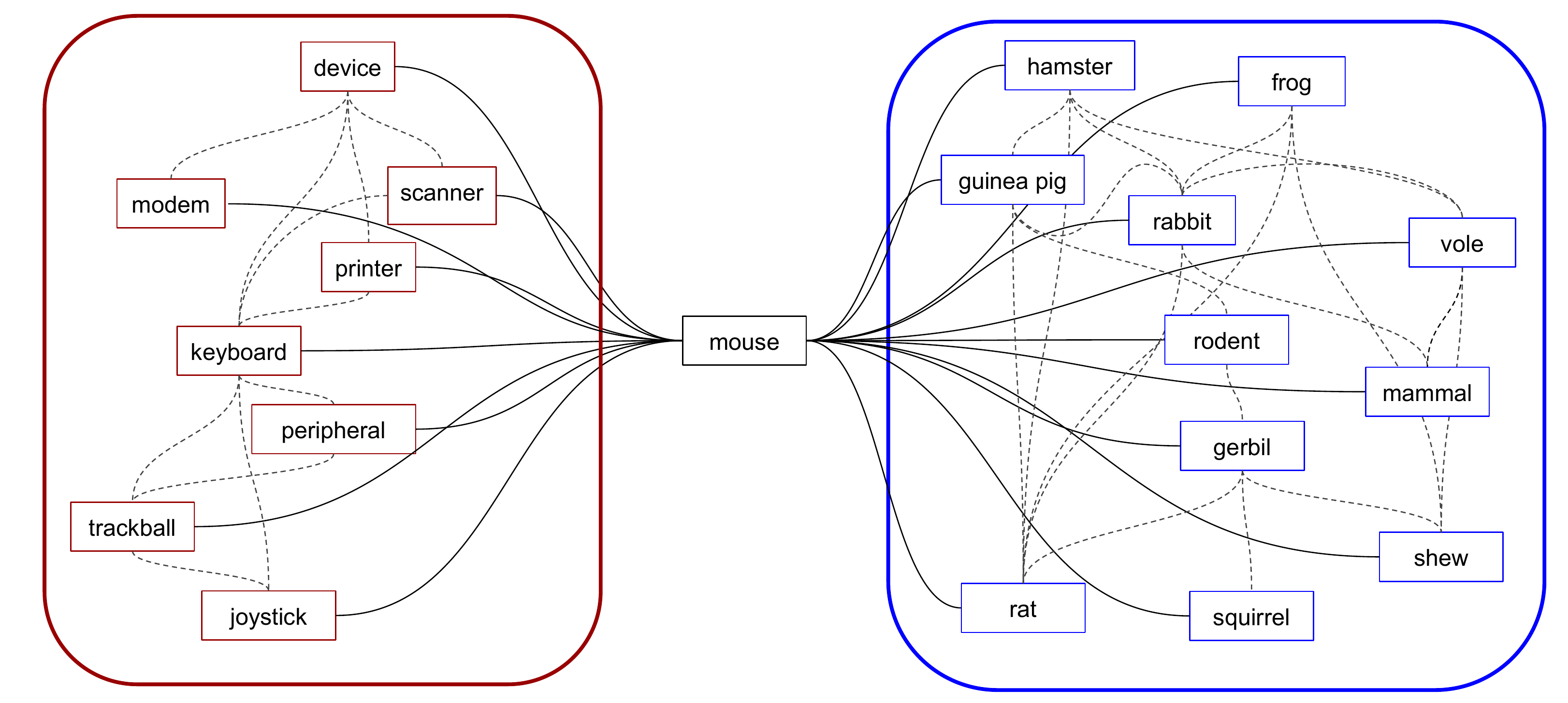}
\end{center}
\caption{Example of graph-based word sense induction for the word \term{mouse}: the two circles denote two induced word senses, as found by analysis of the ego-graph of \term{mouse}.}
\label{fig:interpret-wsd}
\end{figure}

In the next stage, we induce a sense inventory on top of the DT by clustering ego-networks of similar words. In our case, an inventory represents senses by a word cluster, such as \{\term{rat}, \term{rodent}, \term{monkey}, \ldots\} for the `animal' sense of the word \term{mouse}.

Sense induction is conducted one word $t$ at the time on the distributional thesaurus $DT$. First, we retrieve nodes of the ego-network $G$ of $t$ being the $N$ most similar words of $t$ according to the $DT$. Figure~\ref{fig:interpret-wsd} presents a sample ego-network of related words\footnote{See the \url{Serelex.org} system for further visualizations of ego-networks of semantically related words~\cite{panchenko2013serelex}.}. Note that the target word $t$ itself is excluded during clustering. Second, we connect each node in $G$ to its $n$ most similar words according to $DT$. The $n$ parameter regulates the granularity of the induced sense inventory: we experiment with $n \in \{200, 100, 50\}$ and $N = 200$.  Finally, the ego-network is clustered with Chinese Whispers~\cite{Biemann2006}, a nonparametric algorithm that discovers the number of clusters (word senses, in our case) automatically. The algorithm is iterative and proceeds in a bottom-up fashion. Initially all nodes have distinct cluster identifiers. At each step, a node obtains the cluster identifier from the \emph{dominant} cluster in its direct neighborhood, which is the cluster with the highest sum of edge weights to the current node. 

The choice of Chinese Whispers among other algorithms, such as HyperLex~\cite{Veronis2004} or Markov Cluster Algorithm~\cite{van2008graph}, was motivated by the absence of meta-parameters, its state-of-the-art performance on Word Sense Induction tasks~\cite{di2013clustering}, as well as its efficiency (time-linear in the number of edges), see \cite{cecchini17} for a comparative evaluation.

\subsection{Labeling Induced Senses with Hypernyms and Images}
\label{sec:hypernyms}

At the third stage, each sense cluster  is automatically labeled to characterize it in more detail and to improve its interpretability. First, we extract hypernyms from the input corpus. Here, we rely on the Hearst \shortcite{Hearst1992} patterns, yet the approach we use can benefit also from more advanced methods for extraction of hypernyms, e.g.\ HypeNet \cite{shwartz-goldberg-dagan:2016:P16-1} or the Dual Tensor Model \cite{glavas17}. Note that despite their simplicity, Hearst patterns still are a strong baseline, used for applications like, for instance, taxonomy induction~\cite{panchenko2016taxi,bordea2016semeval}.

Second, we rank the quality of a hypernym $h$ to act as generalization for the meaning expressed by cluster $c$ on the basis of the product of two scores, namely \textit{hypernym relevance} and \textit{coverage}:
\begin{align*}
 relevance (c,h) = \sum_{w \in c} rel(t,w) \cdot freq(w,h)\\
 coverage(c,h) = \sum_{w \in c} \min(freq(w,h),1)
\end{align*}
where $rel(t,w)$ is the relatedness of the cluster word $w$ to the target word $t$ (the ambiguous word in an ego-network, cf.\ Figure~\ref{fig:interpret-wsd}), and $freq(w,h)$ is the frequency of the hypernymy relation $(w,h)$ as extracted via patterns. Thus, a highly-ranked hypernym $h$ needs to be observed frequently in a hypernym pattern, but also needs to be confirmed by several cluster words. This stage results in a ranked list of labels that specify the word sense, which we add to the PCZ. The highest-scoring hypernym is further used in the title of the word sense, e.g.\term{mouse (device)} or \term{mouse (animal)}.

Finally, to further improve the interpretability of the induced senses, we add images to our sense clusters as follows. Previous work \cite{faralli-navigli:2012:EMNLP-CoNLL} showed that Web search engines can be used to bootstrap sense-related information. Consequently, we assign an image to each word in the cluster querying the Bing image search API\footnote{\url{https://azure.microsoft.com/en-us/services/cognitive-services/search}} using the query composed of the target word and its highest-scoring hypernym, e.g. \term{mouse device}. The first image result of this query is selected to represent the induced word sense. This step is optional in our pipeline, and is primarily aimed at improving the user interaction with the word sense inventory. 

The resulting sense representation is illustrated in Figure~\ref{fig:demo} for two induced senses of the word \term{mouse}. Providing to the user hypernyms, images, list of related senses and the list of the most salient context clues ensures interpretability of each sense. Note that all these elements are obtained without manual intervention; see \cite{panchenko2017d} for more details. 

\subsection{Disambiguation of Related Terms and Hypernyms}
\label{sec:disambiguation}

\begin{algorithm}[t]
\footnotesize
\caption{Disambiguation of related terms and hypernyms}
\label{alg:closure}
\begin{algorithmic}[1]
\REQUIRE $WSI$, a word sense inventory in the form of a set of tuples $(word$, $sense\_id$, $cluster$, $isas)$, where $cluster$ and $isas$ have no sense identifiers.
\ENSURE $PCZ$, a proto-conceptualization in the form of a set of tuples $(word$, $sense\_id$, $cluster'$, $isas')$, where $cluster'$ and $isas'$ are disambiguated with respect to sense inventory of the $WSI$.
\STATE $PCZ = \emptyset$
\FORALL{$(word,sense\_id, cluster, isas) \in WSI$} 
	\STATE $cluster'$ = \textsc{disambiguate} ($cluster, word, WSI$)
	\STATE $isas'$ = \textsc{disambiguate} ($isas, word, WSI$)

  	\STATE $PCZ = PCZ \cup (word, sense\_id, cluster', isas')$
\ENDFOR  
 \item[]
  \item[]
\textbf{function} \textsc{disambiguate} ($cluster, cword, WSI$)
	\STATE $cluster' = \emptyset$
	\STATE $context = cluster \cup (cword, 1.0)$ 
	\FORALL{$word, sim \in cluster$} 
 		\STATE $sense\_id = -1, max\_sim = 0$
 		\FORALL{$(dword,dsense\_id, dcluster,disas) \in $ \textsc{getSenses}$(word, WSI)$}
  			\IF{$sim(context, dcluster) > max\_sim $} 
  				\STATE $sense\_id = dsense\_id$ 
  				\STATE $max\_sim = sim$ 
    		\ENDIF
  		\ENDFOR 
  		\STATE $cluster' = cluster' \cup (word, sim, sense\_id)$
  	\ENDFOR
\RETURN $cluster'$	
\end{algorithmic}
\end{algorithm}

Next, we disambiguate the lexical graphs induced in the previous step. Each word in the induced inventory has one or more senses: however, the list of related words and hypernyms of each induced sense does not carry sense information yet. In our example (Table \ref{tab:DDTex}) the sense of \term{mouse} for the entry \sense{keyboard:1} could have either referred to the `animal' or the `electronic device'. Consequently, we apply a semantic closure procedure to arrive at a resource in which all terms get assigned a unique, best-fitting sense identifier.  Our method assigns each disambiguation target word $w$  -- namely, a similar or superordinate term from each sense of the induced word sense inventory -- the sense $\hat{s}$ whose context (i.e., the list of similar or superordinate terms) has the maximal similarity with the target word's context (i.e., the other words in the target word's list of similar or superordinate items). We use the cosine similarity between context vectors to find the most appropriate sense $\hat{s}$ matching the ``context'' of an ambiguous word $cluster$ one of its ``definitions'' $WSI(w').cluster$:
\begin{equation}
  \hat{s} = \argmax_{(w',\_,cluster,\_) \in WSI(w)} cos(WSI(w').cluster,cluster).
\label{eq:closure}
\end{equation}
\noindent
This way we are able to link, for instance, \term{keyboard} in the list of similar terms for \sense{mouse:1} to its `device' sense (\sense{keyboard:1}), since \sense{mouse:1} and \sense{keyboard:1} share a large amount of terms from the information technology domain. This simple, local approach is scalable (cf.\ the complexity analysis at the end of this section) and it performs well, as we show later in the evaluation.

Algorithm \ref{alg:closure} presents our method to compute the semantic closure. The input is a JoBimText model as  a set of tuples $(word, sense\_id, cluster, isas)$, where $cluster$ is a list of similar terms in the format $(word_i,sim_i)$ with $sim_i$ being the similarity value between $word$ and $word_i$, and $isas$ is a list of hypernym terms in the same format. The algorithm outputs a proto-conceptualization in the form of a set of tuples $(word, sense\_id, cluster', isas')$, where $cluster'$ is a list of disambiguated similar terms and $isas'$ is a list of disambiguated hypernym terms both in the format $(word_i,sim_i, sense\_id_i)$. The algorithm starts by creating an empty proto-conceptual\-iza\-tion structure $PCZ$. For each entry of an input JoBimText model, we disambiguate related words ($cluster$) and  hypernym terms ($isas$) with the function \textsc{Disambiguate} (lines 3-4). This function retrieves for each $word$ in a $cluster$ the set of its senses with the \textsc{GetSenses} function. 
Next, we calculate similarity between the $cluster$ of the $word$ and the cluster of the candidate sense (denoted as $dcluster$). The $word_i$ obtains the $sense\_id$ of the candidate sense that maximizes this similarity (lines 8-13).

Our disambiguation approach is a rather straightforward algorithm based on similarity computations. Despite its simplicity, we are able to achieve a disambiguation accuracy in the high 90 percent range for noun word senses, while at the same time having a time-linear complexity in the number of word senses, as we will show in the evaluation below (Section \ref{sec:exp1}). We can assume, in fact, that the function \textsc{GetSenses} has a run-time complexity of $O(1)$ and the function $cos$ (Equation \ref{eq:closure}) has complexity $O(m)$, where $m$ is the average number of neighbors of each word sense. Then, the run-time complexity of the algorithm is $O(n*m^2*k)$, where $n$ is the overall number of induced word senses, and $k$ is the average polysemy of a word. Since $k$ is small and $m$ is bound by the maximum number of neighbors (200 in our case), the amortized run-time is linear in the vocabulary size. This makes our approach highly scalable: in recent experiments we have been accordingly able to apply our method at web scale on the CommonCrawl\footnote{\url{https://commoncrawl.org}}, the largest existing public repository of web content.
 
\subsection{Construction of Sense Feature Representations}
\label{sec:sensevec}

Finally, we calculate feature representations for each sense in the induced inventory -- that is, grammatical dependency features that are meant to provide an aggregated representation of the contexts in which a word sense occurs.

We assume that a word sense is a composition of cluster words that represent the sense and accordingly define a sense vector as a function of word vectors representing cluster items.  Let $W$ be a set of all words in the training corpus and let $S_i = \{w_1, \dots, w_n\} \subseteq W$ be a sense cluster obtained in a previous step. Consider a function $\mathit{vec_w}: W \rightarrow \mathbb{R}^m$ that maps words to their vectors and a function $\gamma_i: W \rightarrow \mathbb{R}$ that maps cluster words to their weight in the cluster $S_i$. The sense vector representation (the context clues) is then a weighted average of word vectors:
\begin{equation}
S_i = \frac{\sum_{k=1}^{n} \gamma_i(w_k) \mathit{vec_w}(w_k)  }{ \sum_{k=1}^n \gamma_i(w_k)}.
\end{equation}
Table~\ref{tab:DDTex} (column 4) provides an example of such feature representations. While the averaged word vectors are ambiguous and can contain features related to various senses, features with high weights tend to belong to the target sense as the secondary senses of the averaged words vectors rarely match semantically, hence the aggregation amplifies the correct sense.

This concludes the description of steps we use to construct proto-conceptualizations from text corpora. 

\subsection{Inducing PCZs with Dense Vector Representations}
\label{sec:densevec}

In this section, we briefly describe alternative routes to the construction of a proto-conceptualization from text in an unsupervised way. In the remainder of this article, we will rely on the results of the approach described above. The goal of this section is to show that our overall framework is agnostic to the type of underlying distributional semantic model. In this section, we consider three approaches to generating a PCZ using word or sense embeddings. 

\paragraph{\bf Option 1: Inducing PCZs using word embeddings with explicit disambiguation.} 

\begin{table}
\footnotesize
\begin{tabular}{lcc}
\toprule
 & \bf \parbox{4.8cm}{Sparse vectors (JoBimText)} & \bf \parbox{4.8cm}{Dense vectors (word2vec)} \\ \midrule
mouse (animal) & \parbox{4.8cm}{ rat, rodent,
monkey, pig, animal, human, rabbit, cow}  & \parbox{4.6cm}{ rat, hamster, hedgehog, mole, monkey, kangaroo, skunk } \\ \midrule
mouse (device) & \parbox{4.8cm}{ keyboard, computer, printer, joystick, stylus, modem} & \parbox{4.6cm}{ cursor, keyboard, AltGr, chording, D-pad, button, trackball}  \\\bottomrule
\end{tabular}
\caption{Sense inventories derived from the Wikipedia corpus via a sparse count-based (JoBimText) and dense predict-based (Skip-gram) distributional models. }
\label{tab:embeddings1}
\end{table}

As illustrated in Figure~\ref{fig:workflow}, the first stage of our approach involves the computation of a graph of semantically similar words. Above, the graph was induced using a count-based model, however, any of the models listed in Table~\ref{tab:word2vecf} can be used to generate such a graph of ambiguous words including  models based on dense vector representations, such as the Skip-gram model. In this strategy, one would need to generate top nearest neighbors of word on the basis of cosine similarity between word embeddings. Table~\ref{tab:embeddings1} shows an excerpt of nearest neighbors generated using the JoBimText and word2vec toolkits. The obtained word graphs can be subsequently used to induce word senses using the graph-based approach described in Section~\ref{sec:wsi}.  The obtained clusters can be labeled using a database of hypernyms exactly in the same way as for the models based on the count-based JoBimText framework (cf.\ Section~\ref{sec:hypernyms}). All further steps of the workflow presented in Figure~\ref{fig:workflow} remain the same.

The main difference between the approach described above and the methods based on dense representations of words is the representation of the context clues of PCZ (cf. Table~\ref{tab:DDTex}). In the case of an underlying sparse count-based representation, context clues remain human-readable and interpretable, whereas in case of dense representations, context clues are represented by a dense vector embedding, and it is not straightforward to aggregate context clues over sense clusters.

\begin{table}
\footnotesize
\centering
\begin{tabular}{ll}
\toprule
\bf Vector & \bf \parbox{5.7cm}{Nearest Neighbors} \\ \midrule
 mouse & \parbox{11cm}{rat, keyboard, hamster, hedgehog, monkey, kangaroo, cursor,  button  }\\ \midrule
  mouse:0 & \parbox{11cm}{ rat:0, hamster:0, hedgehog:1, mole:0, monkey:0, kangaroo:0, skunk:0} \\ \midrule
   mouse:1 & \parbox{11cm}{ cursor:0, keyboard:1, AltGr:0, chording:1, D-pad:0, button:0 } \\ 
\bottomrule
\end{tabular}
\caption{A Skip-gram based PCZ model by (Pelevina et al. 2016): Neighbors of the word \term{mouse} and the induced senses. The neighbors of the initial vector belong to both senses, while those of sense vectors are sense-specific.}
\label{tab:embeddings2}
\end{table}

\paragraph{\bf Option 2: Inducing PCZs using word embeddings without explicit disambiguation.} 

The first two stages of this approach are the same compared to the previous strategy. Namely, first one needs to generate a graph of ambiguous semantically related words (Section~\ref{sec:dt}) and then to run ego-network clustering to induce word senses (Section~\ref{sec:wsi}). However, instead of explicit disambiguation of nearest neighbors (Section~\ref{sec:disambiguation}), the third stage could obtain vector sense representations by averaging the word embeddings of sense clusters (Section~\ref{sec:sensevec}). Finally, disambiguated nearest neighbors can be obtained by calculating nearest neighbors of each sense vector in the space of word sense embeddings. This step is equivalent to the computation of a distributional thesaurus (Section~\ref{sec:dt}), however it directly yields  disambiguated nearest neighbors (cf. Table~\ref{tab:embeddings2}). Note however that, disambiguation of hypernyms using Algorithm \ref{alg:closure} is still required when using this approach. 

This approach was explored in our previous work \cite{pelevina-EtAl:2016:RepL4NLP}, where we showed that words sense embeddings obtained in this way can be successfully used for unsupervised word sense disambiguation, yielding results comparable to the state of the art. 

\paragraph{\bf Option 3: Inducing PCZ using word sense embeddings.} 

Finally, a PCZ can be also induced using sparse~\cite{reisinger2010multi} and dense~\cite{neelakantan-EtAl:2014:EMNLP2014,li2015multi,bartunov2015breaking} multi-prototype vector space models (the latter are also known as word sense embeddings). These models directly induce sense vectors from a text corpus, not requiring the word sense induction step of our method (Section~\ref{sec:wsi}). Instead of ego-network-based sense induction, these methods rely on some form of context clustering, maintaining several vector representations for each word type during training. To construct a PCZ using such models within our framework, we need to compute a list of nearest neighbors (Section~\ref{sec:dt}), label the obtained sense clusters with hypernyms (Section~\ref{sec:hypernyms}) and disambiguate these hypernyms using Algorithm 1. The sense vectors replace the aggregated context clues, so the stage described in Section~\ref{sec:sensevec} is superfluous for this option as well. 

We also experimented in previous work with the construction of proto-conceptualizations using this approach \cite{panchenko16}, showing how to use sense embeddings for building PCZs, reaching satisfactory levels of recall and precision of matching as compared to a mapping defined by human judges. 

While an empirical comparison of these options would be interesting, it is beyond the scope of this paper, where our main point is to demonstrate the benefits of linking manually created lexical resources with models induced by distributional semantics (by example of a count-based model). 

\section{Linking a Proto-conceptualization to a Lexical Semantic Resource}
\label{sec:linking}

This section describes how a corpus-induced semantic network (a proto-conceptualization) is linked to a manually created semantic network, represented by a lexical resource. 

\subsection{Linking Induced Senses to Senses of the Lexical Resource}
\label{subsec:linking}

\begin{algorithm}[t]

\caption{Linking induced senses to senses of a lexical resource}\label{alg:linking}
\begin{algorithmic}[1]
\footnotesize
\REQUIRE $T = \{(j_i,R_{j_i},H_{j_i})\}$, $W$, $th$, $m$\\ 
\ENSURE $M={(source,target)}$
\STATE $M = \emptyset$ \label{alg:init}
\FORALL{$(j_i,R_{j_i},H_{j_i}) \in T.monosemousSenses$} \label{alg:monoinit}
  \STATE $C(j_i)=W.getSenses(j_i.lemma,j_i.POS)$
  \IF{$|C(j_i)|==1$, let $C(j_i)=\{c_0\}$} \label{alg:monoexists}
    \IF{ $sim(j_i,c_0,\emptyset) \geq th$} \label{alg:similarity}
      \STATE $M = M \cup \{(j_i,c_0)\}$
    \ENDIF
  \ENDIF
\ENDFOR \label{alg:monoend}
\FOR{$step=1$; $step \leq m$ ; $step=step+1$} \label{alg:startiter}
\STATE $M_{step} = \emptyset$ \label{alg:initstep}
  \FORALL{$(j_i,R_{j_i},H_{j_i}) \in T.senses/M.senses$}
    \STATE $C(j_i)=W.getSenses(j_i.lemma,j_i.POS)$
    \FORALL{$c_k \in C(j_i)$} \label{alg:brank}
      \STATE $rank(c_k)=sim(j_i,c_k,M)$
    \ENDFOR \label{alg:erank}
    \IF{$rank(c_k)$ has a single top value for $c_t$}  \label{alg:bselect}
	\IF{ $rank(c_t) \geq th$} \label{alg:th}
	    \STATE $M_{step} = M_{step} \cup \{(j_i,c_t)\}$
	\ENDIF
    \ENDIF \label{alg:eselect}
\ENDFOR
      \STATE $M = M \cup M_{step}$ \label{alg:collect}
\ENDFOR\label{alg:enditer}

\FORALL{$(j_i,R_{j_i},H_{j_i}) \in T.senses/M.senses$} \label{alg:bnovel}
      \STATE $M = M \cup \{(j_i,j_i)\}$
\ENDFOR \label{alg:enovel}
\RETURN $M$
\end{algorithmic}

\end{algorithm}

Now, we link each sense in our proto-conceptual\-iza\-tion (PCZ) to the most suitable sense (if any) of a Lexical Resource (LR, see Figure \ref{fig:workflow} step 3). There exist many algorithms for knowledge base linking \cite{pavel13}: here, we build upon simple, yet high-performing previous approaches to linking LRs that achieved state-of-the-art performance. These rely at their core on computing the overlap  between the bags of words built from the LRs' concept lexicalizations, e.g., \cite{navigli12,gurevych12} (\emph{inter alia}). Specifically, we develop i) an iterative approach -- so that the linking can benefit from the availability of linked senses from previous iterations -- ii) leveraging the lexical content of the source and target resources. Algorithm \ref{alg:linking} takes as input:
\begin{enumerate}[leftmargin=4mm]
\item a PCZ $T = \{(j_i,R_{j_i},H_{j_i})\}$ where $j_i$ is a sense identifier (i.e. \term{mouse:1}), $R_{j_i}$ the set of its semantically related senses  (i.e. $R_{j_i}=\{$\term{keyboard:1}, \term{computer:0}, \dots $\}$ and $H_{j_i}$ the set of its hypernym senses (i.e. $H_{j_i}=\{$\term{equipment:3}, \dots $\}$;
\item a LR $W$: we experiment with: WordNet, a lexical database for English and BabelNet, a very large multilingual `encyclopedic dictionary';
\item a threshold $th$ over the similarity between pairs of concepts and a number $m$ of iterations as a stopping criterion.
\end{enumerate}
The algorithm outputs a mapping $M$, which consists of a set of pairs of the kind ${(source,target)}$ where $source \in T.senses$ is a sense of the input PCZ $T$ and $target \in W.senses \cup {source}$
is the most suitable  sense of $W$ or $source$ when no such sense has been identified.

The algorithm starts by creating an empty mapping $M$ (line \ref{alg:init}). Then for each monosemous sense (e.g., \term{Einstein:0} is the only sense in the PCZ for the term \term{Einstein}) it searches for a candidate monosemous sense (lines \ref{alg:monoinit}-\ref{alg:monoend}). If such monosemous candidate senses exist (line \ref{alg:monoexists}), we compare the two senses (line \ref{alg:similarity}) with the following similarity function: 
\begin{equation}
sim(j,c,M)=\frac{|T.BoW(j,M,W) \cap W.BoW(c)|}{|T.BoW(j,M,W)|}, 
\label{formula:smilarity}
\end{equation}
\noindent
where
\begin{enumerate}[leftmargin=4mm]
\item{$T.BoW(j,M,W)$} is the set of words containing all the terms extracted from related/hypernym senses of $j$ and all the terms extracted from the related/hypernym (i.e., already linked in $M$) synsets in W. For each synset from the LR, we use all synonyms and content words of the gloss.
\item{$W.BoW(c)$} contains the synonyms and the gloss content words for the synset $c$ and all the related synsets of $c$.   
\end{enumerate}
Then a new link pair $(j_i,c_0)$ is added to $M$ if the similarity score between $j_i$ and $c_0$ meets or exceeds the threshold $th$ (line \ref{alg:similarity}). 
At this point, we collected a first set of disambiguated (monosemous) senses in $M$ and start to iteratively disambiguate the remaining (polysemous) senses in $T$ (lines \ref{alg:startiter}-\ref{alg:enditer}).
This iterative disambiguation process is similar to the one we described for the monosemous case (lines \ref{alg:monoinit}-\ref{alg:monoend}), with the main difference that, due to the polysemy of the candidates synsets, we instead use the similarity function to rank all candidate senses (lines \ref{alg:brank}-\ref{alg:erank}) and select the top-ranked candidates for the mapping (lines \ref{alg:bselect}-\ref{alg:eselect}). At the end of each iteration, we add all collected pairs to $M$ (line \ref{alg:collect}). Finally, all unlinked $j$ of $T$, i.e.\ induced senses that have no corresponding LR sense, are added to the mapping $M$ (lines \ref{alg:bnovel}-\ref{alg:enovel}).

\paragraph{\textbf{Comparison with other mapping algorithms.}} Previous work for the construction of BabelNet \cite{navigli12} and UBY \cite{gurevych12}  looked at the related problem of matching heterogeneous lexical semantic resources, i.e., Wikipedia and WordNet. In our scenario, however, we aim instead at establishing a bridge between any of these latter reference KBs and a proto-conceptualization -- i.e., a fully disambiguated distributional semantic representation of distributionally induced word senses (cf.\ Section \ref{sec:sensevec}). Since we are working with a PCZ on the source side, as opposed to using Wikipedia, we cannot rely on graph-based algorithms such as PageRank \cite{niemann11} `out of the box':  while PCZs can be viewed as graphs, these are inherently noisy and require cleaning techniques in order to remove cycles and wrong relations (cf.\ Section \ref{sec:applications} where we accordingly address the topic of taxonomy induction and cleaning within our framework). Similarly, the fact that PCZs are automatically induced from text -- and hence potentially noisier that clean collaboratively generated content from Wikipedia -- forces us to limit evidence for generating the mapping to local information, as opposed to, e.g., graph-based expansions used to boost the recall of BabelNet-WordNet mappings, cf. \cite{navigli12}. To overcome this `locality' constraint, we develop an iterative approach to indirectly include non-local evidence based on previous mapping decisions. Our algorithm, in fact, uses previous mappings to additionally expand the bag-of-word of the candidate PCZ sense to be mapped, based on related/hypernym synsets linked in the previous iterations only (i.e., to keep the expansion `safe', cf.\ Equation \ref{formula:smilarity}).

\subsection{Typing of the Unmapped Induced Senses}
\label{subsec:typing}

\begin{algorithm}[t]
\footnotesize

\caption{Typing of the unmapped induced senses}\label{alg:suggesting}
\begin{algorithmic}[1]
\REQUIRE $M={(source,target)}$, $W$\\ 
\ENSURE $H={(source,type)}$
\STATE $H = \emptyset$ \label{alg:ts_init}
\FORALL{$(source,target) \in M$} \label{alg:ts_in1}
  \IF{$target \notin W$} \label{alg:ts_nomapped}
      \STATE $Rank=0$
	\FORALL{$related \in R_{source}, \exists (related,trelated) \in M, trelated \in W$ } \label{alg:ts_in2}
	  \FORALL{$hop \in (1,2,3)$} 
	    \FORALL{$ancestor \in W.ancestors(trelated,hop)$} 
	       \STATE $Rank(ancestor)=Rank(ancestor)+1.0/hop$
	      \ENDFOR     
	   \ENDFOR  
	\ENDFOR  \label{alg:ts_end2}
      \FORALL{$ntype \in Rank.top(top_h)$} \label{alg:ts_in}
	\STATE $H=H \cup (source,ntype)$
      \ENDFOR  
 \ENDIF
 \ENDFOR \label{alg:ts_end1}
\RETURN $H$
\end{algorithmic}
\end{algorithm}

An approach based on the bag-of-words from concept lexicalizations has the advantage of being simple, as well as high performing as we show later in the evaluation -- cf.\ also findings from \cite{navigli12}. However, there could be still PCZ senses that cannot be mapped to the target lexical resource, e.g., because of vocabulary mismatches, sparse concepts' lexicalizations, or because they are simply absent in the resource.

Consequently, in the last phase of our resource creation pipeline we link these `orphan' PCZ senses (i.e., those from lines \ref{alg:bnovel}-\ref{alg:enovel} of Algorithm \ref{alg:linking}), in order to obtain a unified resource, and propose a method to infer the type of those concepts that were not linked to the target lexical resource. For example, so far we were not able to find a BabelNet sense for the PCZ item \sense{Roddenberry:10} (the author of `Star Trek').  However, by looking at the linked related concepts that share the same BabelNet hypernym -- e.g.\ the PCZ items \sense{Asimov:3} \emph{is-a} \sense{author$_{\textrm{\emph{BabelNet}}}$}, \sense{Tolkien:7} \emph{is-a} \sense{author$_{\textrm{\emph{BabelNet}}}$}, \sense{Heinlein:8} \emph{is-a} \sense{author$_{\textrm{\emph{BabelNet}}}$}, etc.\ -- we can infer that \sense{Roddenberry:10} \emph{is-a} \sense{author:1}, since the latter was linked to the Babel synset \sense{author$_{\textrm{\emph{BabelNet}}}$}.

The input of Algorithm \ref{alg:suggesting} consist of the mapping $M$ of a PCZ to a lexical resource $W$ (cf.\ Algorithm \ref{alg:linking}). The output is a new mapping $H$ containing pairs of the kind $(source,type)$ where $type$ is a type in $W$ for the concept $source \in PCZ$.
We first initialize the new mapping $H$ as an empty set (line \ref{alg:ts_init}). Then for all the pairs $(source,target)$ where the target is a concept not included in the target lexical resource $W$ (line \ref{alg:ts_nomapped}), we compute a rank of all the ancestors of each related sense that has a counterpart $trelated$ in $W$ (lines \ref{alg:ts_in2}-\ref{alg:ts_end2}). In other words, starting from linked related senses $trelated$, we traverse the taxonomy hierarchy (at most for 3 hops) in $W$ and each time we encounter a sense $ancestor$ we increment its rank by the inverse of the distance to $trelated$. Finally we add the pairs $(source,ntype)$ to $H$ for all the $ntype$ at the top $top_h$ in the $Rank$.

Finally, our final resource consists of: i) the proto-conceptualization (PCZ); ii) the mapping $M$ of PCZ entries to the lexical resource (e.g., WordNet or BabelNet); iii) the mapping $H$ of suggested types for the PCZ entries not mapped in $M$.

\section{Experiments}
\label{sec:experiments}

In this section, we present results of four experiments, which intrinsically and extrinsically evaluate the quality of our hybrid aligned resource. 

\subsection{Corpora Used for the Induction of Proto-conceptualizations}
\label{sec:datasets}

\begin{table}
\scriptsize
\centering

\begin{tabular}{lllll} 
 \toprule 
\bf Name & \bf Language & \bf Source of texts & \bf Genre & \bf Size \\ \midrule
wiki & English & Text of Wikipedia articles & Encyclopedic & 35 million sent. \\
news & English & News articles: Gigaword and LCC & Narrative, publicistic & 105 million sent. \\
\bottomrule

\end{tabular}
\caption{Text corpora used in our experiments to induce distributional disambiguated semantic networks (proto-conceptualizations, PCZs).}
\label{tbl:corpora}
\end{table}

We evaluate our method using texts of different genres, namely standard newswire text vs.\ encyclopedic texts in order to examine performance in different settings. The corpora, described in Table~\ref{tbl:corpora}, are a 105 million sentence news corpus composed of Gigaword \cite{Gigaword} and the Leipzig Corpora Collection (LCC) \cite{LCC}\footnote{\url{http://corpora.uni-leipzig.de}}, and  a 35 million-sentence Wikipedia corpus\footnote{The \textit{wiki} corpus is downloadable (cf. Section~\ref{sec:conclusions}). The \textit{news} corpus is available by request due to license restrictions of the Gigaword corpus.} from a 2011 dump.

We opt for these text collections because they were previously extensively used for the evaluation of distributional models based on the JoBimText framework \cite{BiemannRiedl2013,Riedl13scale}. Specifically, previous work \cite{Riedl13scale} experimented with the induction of distributional models on the basis of both corpora, and showed that the quality of semantic similarity (which, in turn, is used to build the distributional thesaurus, cf.\ Section \ref{sec:dt}) increases with corpus size.
%
%
%
Since `more data' helps, we experiment in this work with the full-sized corpora. Further description of the \textit{wiki} and \textit{news} text collections can be found in \cite{Riedl13scale} and (Riedl 2016, p.\ 94).

We experiment with different parameterizations of the sense induction algorithm to obtain proto-conceptualizations (PCZ) with different average sense granularities, since \emph{a priori}, there is no clear evidence for what the `right' sense granularity of a sense inventory should be. Chinese Whispers sense clustering with the default parameters ($n=200$) produced an average number of 2.3 (news) and 1.8 (wiki) senses per word with the usual power-law distribution of sense cluster sizes. Decreasing connectivity of the ego-network via the $n$ parameter leads to more fine-grained inventories  (cf. Table~\ref{table:datasetsstatistics}).  

\begin{figure}[t!]
 \centering
 \includegraphics[width=0.9\columnwidth]{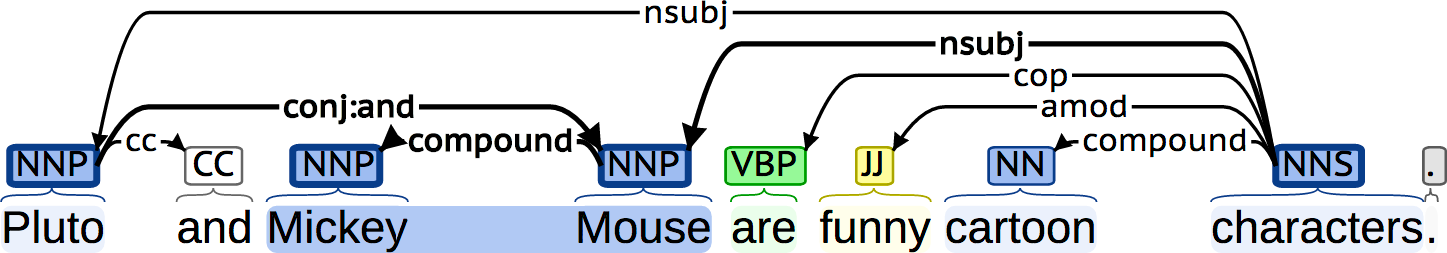}
 \caption{Extraction of distributional dependency features for a multiword expression \term{Mickey Mouse}: all outgoing dependencies are used as features. This image was created using the Stanford dependency visualizer (\texttt{http://nlp.stanford.edu:8080/corenlp}).}
 \label{fig:mwe}
\end{figure}

Finally, we use the method described in \cite{RiedlBiemann2015} to compute a dataset that includes automatically extracted multiword terms using the Wikipedia corpus (\textit{wiki-p1.6-mwe}). Since most of the multiwords are monosemous, average polysemy of this dataset decreased from 1.8 to 1.6 for the analogous model without multiwords (\textit{wiki-p1.8}). To obtain a feature representation of a multiword expression, we gathered all outgoing dependency relations of this term as illustrated in Figure~\ref{fig:mwe}.

\begin{table}
\scriptsize	 
 \centering
 \begin{tabular}{lrrrrrrrrrrr} 
 \toprule 
&   & \multicolumn{3}{l}{\bf Words} &\multicolumn{1}{l}{\bf Senses} & \multicolumn{2}{l}{\bf Polysemy} & \multicolumn{2}{l}{\bf Rel.senses} &  \multicolumn{2}{l}{\bf Hyper.} \\
\bf PCZ & $n$  & \# &mono & poly & \# & avg. & max & \# & avg. & \# & avg.\\
 \midrule
 news-p1.6& 200 & 207k  & 137k &  69k &  332k & 1.6 & 18 & 234k & 63.9 & 15k & 6.9 \\ 
 news-p2.3& 50&  200k &  99k  & 101k & 461k & 2.3 & 17 & 298k & 44.3 & 15k  & 5.8 \\ 
 wiki-p1.8& 200 & 206k & 120k  & 86k & 368k  & 1.8  & 15 & 300k & 59.3 & 15k & 4.4 \\ 
 wiki-p6.0& 30& 258k & 44k  & 213k& 1.5M & 6.0 & 36 & 811k & 16.9 & 52k & 1.7 \\ 
 wiki-p1.6-mwe & 200& 465k & 288k  & 176k& 765k& 1.6 & 13 & 662k & 46.6 & 30k & 3.2 \\
 \bottomrule
 \end{tabular}
\caption{Structural analysis of our five word sense inventories of the proto-conceptualizations (PCZs) used in our experiments.}
\label{table:datasetsstatistics}
 \end{table}

In Table \ref{table:datasetsstatistics}, we present statistics for the five different resources we induce from our corpora. For each dataset, we report the counts of overall number of words (vocabulary size), including monosemous words and polysemous ones, respectively. For each PCZ we report the cardinality, the average polysemy and the maximum polysemy. Finally, we report the overall and the average number of related senses and hypernyms. Numbers vary across datasets due to the different nature of the two source corpora and the selection of different parameter values for sense induction.

While inducing senses directly from corpus data allows for large coverage and flexibility, it also makes it difficult to evaluate the quality of the resulting sense clusters \cite{agirre07}. Since  we do not a priori know the sense granularity of the PCZs, the sense inventory input to our disambiguation and linking algorithms cannot be fixed in advance, e.g., in order to produce a static gold standard. Therefore, in our intrinsic evaluations (Section \ref{sec:exp1}--\ref{sec:exp3}) we assess the quality of our resources by manually validating a sample of the output of the different steps of our method. 
Later in Section \ref{sec:semeval}, we perform an extrinsic evaluation against a gold standard on a WSD benchmarking dataset from a SemEval task.

The PCZ described above were subsequently linked to WordNet 3.1 and BabelNet 2.5 using the method described above. All the models described above as well as the induced sense inventories and word similarity graphs can be accessed online (cf. Section~\ref{sec:conclusions}).

\subsection{Experiment 1: Quality of Disambiguation of the Related Terms}
\label{sec:exp1}

\paragraph{\textbf{Experimental setting.}} 
In this experiment, we evaluate the quality of Algorithm \ref{alg:closure} for the disambiguation of related words (cf. Table~\ref{tab:DDTex}) by performing a post-hoc evaluation using manual judgments on a sample of sense-disambiguated terms from one of our PCZ resources. 
%

We manually selected a set of frequent nouns and proper nouns, such that each word has least two homonymous (as opposed to polysemous) word senses. We deliberately avoided words with polysemous senses, as word sense induction algorithms are known to robustly extract mostly coarse-grained inventories of homonymous words~\cite{di2013clustering,cecchini17}. The words were selected according to two criteria. First, each of the two homonymous word senses should have a comparable frequency -- compare, for instance, the fairly common senses of  \sense{python (animal)} and \sense{python (language)}, as opposed to \sense{boa (animal)} and \sense{boa (language)}, where the `language' sense of the word \term{boa} is much rarer as compared to its `animal' sense. Second, each of these senses should be common enough to be recognizable without the need of consulting a dictionary by a non-native, graduate-level speaker of English. We tested for sense frequencies and popularity by checking that selected senses were found among the top ones as listed in BabelNet. Using these criteria, we manually selected a set of 17 nouns, such as \term{apple}, \term{java}, \term{python}, etc.\footnote{The full list of words, where senses are denoted in the brackets using the respective hypernyms: \term{apache} (tribe$|$software), \term{apple} (fruit$|$company), \term{bank} (river$|$institution), \term{commercial} (ad$|$business), \term{corvette} (car$|$ship), \term{jaguar} (animal$|$car), \term{java} (island$|$technology), \term{lotus} (flower$|$car), \term{mustang} (horse$|$car), \term{pascal} (person$|$language), \term{port} (sea-related$|$computer-related), \term{puma} (animal$|$brand), \term{python} (snake$|$language), \term{ruby} (gem$|$language), \term{sun} (star$|$company), \term{tiger} (animal$|$tank), and \term{viper} (snake$|$car).}

\begin{table}
\footnotesize
\centering
\begin{tabular}{lcccc} 
 \toprule 
\bf Part of Speech  & \bf \# Word Forms & \bf \# Senses & \bf \# Contexts & \bf Accuracy \\ \midrule
Nouns & 15 & 30 & 1055 & 0.94 \\
Proper nouns & 17 & 49 & 1177 & 0.85 \\
Adjectives & 6 & 6 & 566 & 0.63 \\
Verbs & 4 & 6 & 86 & 0.76 \\
\midrule
All & 42 & 91 & 2284 & 0.84 \\
\bottomrule
\end{tabular}
\caption{Accuracy of Algorithm \ref{alg:closure} for disambiguation of related words evaluated on a set of 17 frequent words each having two non-marginal homonymous word senses, e.g.\ as in \sense{mouse (keyboard)} and \sense{mouse (animal)}. 
}
\label{tbl:exp1results}
\end{table}

\begin{table}
\footnotesize
\centering
\begin{tabular}{lccc} 
 \toprule 
\bf Word & \bf Sense ``Definition'' & \bf Sense ``Context'' & \bf Related \\
\midrule
java & \parbox{4.5cm}{UNIX, Linux, Symbian, Unix, OS, Android, Mobile, Solaris, MS-DOS, Windows, iOS} & \parbox{4.5cm}{screen, I/O, multiprocessor, IDE, repository, pak, Blu-ray, Graphics, Video, Itanium, ...} & Yes \\ \midrule

python & \parbox{4.5cm}{hamster, lemurs, turtle, constrictor, lizard, orca, rhinoceros, cobra, ...} & \parbox{4.5cm}{turtle, breeds, cattle, breed, cobra, Bulbul, Kingfisher, Mammals, starling, ...} & Yes \\ \midrule

sun & \parbox{4.5cm}{Hearth, mirror, orb, soil, star, spotlight, temperature, water, solstice, burst, ...} & \parbox{4.5cm}{eel, brave, Celtics, Wrangler, rockies, Chargers, Expos, Cavaliers, Cougars, padre, ...} & No \\ \midrule

tiger & \parbox{4.5cm}{Macaque, deer, rhinoceros, Falcon, mascot, Whale, Gibbon, Hyena, boar, deer, ...} & \parbox{4.5cm}{Nighthawk, Cessna, F-16, Valiant, Corsair, Maurer, Mirage, Reaper, Scorpion, ...} & No \\ 
\bottomrule
\end{tabular}
\caption{Examples provided to the assessor participating in the study with correct judgments. The subject was asked to determine if the first sense cluster (representing a sense definition of the ambiguous related word) is semantically related to the second sense cluster (representing the context of the ambiguous related word). }
\label{tbl:exp1annotion}
\end{table}

Since our resources only partially overlap in terms of sense inventory, and there is no \emph{a priori} reference sense granularity, we cannot perform evaluation on a shared gold standard. Consequently, we opt instead for a post-hoc evaluation of the accuracy of the disambiguation step, namely the fraction of correctly disambiguated related words among all disambiguated words. Post-hoc validations have major limitations in that they are time-consuming, do not scale and hinder direct comparability across methods -- nevertheless, they are commonly used in the field of knowledge acquisition to estimate the quality of knowledge resources \cite{banko07,Suchaneketal:08,carlson10,velardi13} (\emph{inter alia}).

We performed manual validation as follows. We first collected all disambiguated entries of the \textit{wiki-p1.6} model (cf.\ Table~\ref{tab:DDTex}), where these 17 target words appear and randomly sampled 15\% of these entries to make annotation feasible, resulting in a dataset of 2\,884 ambiguous related words in context. We restrict evaluation to the \textit{wiki-p1.6} model for two reasons: an encyclopedic source is expected to provide better sense coverage \emph{a priori}, thus providing us with more evaluation instances, while a low number of clusters is in line with findings that graph-based sense induction methods can produce rather coarse high-quality clusterings \cite{cecchini17}.

Table~\ref{tbl:exp1results} presents statistics of our dataset: note that we gathered word senses of all parts of speech that correspond to the selected 17 words, including verbs and adjectives, for the sake of completeness of our study.  An annotator with previous experience in lexicographic annotation performed the judgment of the 2\,884 contexts in a curated way (using several rounds of feedback on random samples of annotations). The annotator was presented with a table containing four columns: i) the target word,  ii) a sense cluster defining the sense of the target word, iii) a sense cluster that defines the context of the target word. The last column collected the binary answer on the question whether  the `definition' cluster is semantically compatible with the `context' cluster. Table~\ref{tbl:exp1annotion} illustrates two examples of semantically compatible and incompatible clusters. The reasons of incompatibility of sense clusters are either the absence of obvious semantic relations between the words in the clusters (cf.\ the `planet' vs.\ `basketball' sense of \term{sun}) or simply incoherence of one or both sense clusters -- i.e., the case when the annotator cannot figure out the meaning of the sense denoted by a cluster, such as the case for the context cluster of \term{tiger}. The annotator was instructed to consider a sense cluster to be interpretable if it was possible to grasp a dominant meaning by looking at the top 20 words, while allowing for a small fraction of spurious terms (since sense clusters are automatically generated).

\paragraph{\textbf{Results and discussion.}} The results of the experiment are summarized in Table~\ref{tbl:exp1results}.\footnote{The judgments are available for download (cf. Section~\ref{sec:conclusions}).} Performance of the disambiguation procedure for the proper names and nouns ranges from 0.85 to 0.94, thus indicating an overall high quality of the procedure. Note that the word senses of adjectives and verbs  are mostly the result of part-of-speech tagging errors, since in the seed set of 17 words, we added only nouns and proper nouns.  Wrongly tagged words have in general more noisy, un-interpretable clusters.

To better understand the amount of spurious items in our sense clusters, we performed an additional manual evaluation where, for a sample of 100 randomly sampled noun PCZ items, we counted the ratio between wrong (e.g., \term{rat} for the computer sense of \term{mouse}) and correct (\term{keyboard}, \term{computer}, etc.) related sense that were found within the PCZs. We obtained a macro average of 0.0495 and a micro average of 0.0385 wrongly related senses within the PCZs. Moreover, 83\% of the above sample has no unrelated senses, and only 2\% have only a single unrelated sense with a macro average ratio between the wrong and correct related PCZs of 0.067. This indicates that, overall, the amount of spurious senses within clusters is indeed small, thus providing a high-quality context for an accurate disambiguation of noun DT clusters.

\subsection{Experiment 2: Linking Induced Senses to Lexical Resources}
\label{sec:exp2}

\paragraph{\textbf{Experimental setting.}} In this experiment, we evaluate the performance of our linking component (Section \ref{subsec:linking}). For this, we choose two lexical-semantic networks: WordNet \cite{Fellbaum:98}, which has a high coverage on English common nouns, verbs and adjectives, and BabelNet \cite{navigli12}, which also includes a large amount of proper nouns and senses gathered from multiple other sources, including Wikipedia.

We follow standard practices, e.g., \cite{navigli12}, and create five evaluation test sets, one for each dataset from Section \ref{sec:datasets}, by randomly selecting a subset of 300 induced word senses for each dataset, and manually establishing a mapping from these senses to WordNet and BabelNet senses (senses that cannot be mapped are labeled as such in the gold standard).

We compare against two most frequent sense (MFS) baselines, which select from all the possible senses for a given term $t$: 
\begin{enumerate}[leftmargin=4mm]

 \item the most frequent sense in WordNet, where frequencies of senses are observed on a manually annotated semantic concordance \cite{Miller:1993:SC:1075671.1075742}.
 
 \item the most frequent sense in BabelNet. Since BabelNet combines WordNet and Wikipedia, this amounts to: i) the WordNet MFS for senses originally found in WordNet, and ii) the most cited (i.e., internally hyperlinked) Wikipedia page for senses derived from Wikipedia.

\end{enumerate}
The quality and correctness of the mapping is estimated as accuracy on the ground-truth judgments, namely the amount of true mapping decisions among the total number of (potentially, empty) mappings in the gold standard. Each pair $(j,c)$ in a mapping $M$ created with Algorithm \ref{alg:linking} is evaluated as: i) true positive ($TP$) when $c$ is the most suitable sense in the lexical resource for the induced word sense $j$; ii) true negative ($TN$) when $c$ refers to $j$ itself and there are no senses in the lexical resource to capture the meaning expressed by $j$; iii) false positive ($FP$) when $c$ is not the most suitable sense in the lexical resource for the sense  $t$; iv) false negative ($FN$) when $c$ refers to $j$ itself and there is a sense in the lexical resource that captures the same meaning of $j$.

We also evaluate our mapping by quantifying coverage and extra-coverage on the reference resource:
\begin{equation}
  \label{equation:coverage}
  Coverage(A,B)=\frac{|A\cap B|}{|B|}  \qquad     ExtraCoverage(A,B)=\frac{|A/B|}{|B|}
\end{equation}
where $A$ is the set of lexical resource synsets or induced word senses mapped in $M$ using Algorithm \ref{alg:linking}, and $B$ is the whole set of lexical resource synsets. That is, Coverage indicates the percentage of senses of the lexical resource sense inventory covered by the mapping $M$, whereas ExtraCoverage indicates the ratio of senses in $M$ not linked to the lexical resource sense inventory over the total number of senses in a lexical resource. That is, ExtraCoverage is a measure of novelty to quantify the amount of senses discovered in $T$ and not represented by the lexical resource: it indicates the amount of `added' knowledge we gain with our resource based on the amount of senses that cannot be mapped and are thus included as novel senses.


\begin{table}[t]
\footnotesize
\centering
\scalebox{1.0}{\begin{tabular}{lrcccc}
 \toprule
  \multicolumn{6}{c}{\bf WordNet-linked}\\ 
  PCZ &  \#linked senses & Cov. & ExtraCov. & Accuracy & MFS baseline \\ 
  \cline{2-6}\\
  news-p1.6 & 88k & 34.5\% & 206.0\%& 86.9\%& 85.5\% \\
  news-p2.3 & 145k & 38.2\% & 267.0\%& 93.3\% & 85.0\% \\
  wiki-p1.8 & 91k & 35.9\%& 234.7\%& 94.8\% & 80.5\% \\
  wiki-p6.0 & 400k & 49.9\% & 919.9\% & 93.5\% &74.2\%\\
  wiki-mw-p1.6 & 81k & 30.7\% & 581.2\% & 95.3\% &89.7\%\\
 \midrule
 &&&&&\\
 \multicolumn{6}{c}{\bf BabelNet-linked}\\ 
  &  \# linked senses & Cov. & ExtraCov. & Accuracy & MFS baseline \\ 
  \cline{2-6}\\
  news-p1.6 & 164k& 1.3\%& 2.9\%& 81.8\%& 52.3\%\\
  news-p2.3 & 236k & 1.4\%& 3.9\%& 85.1\%& 57.2\%\\
  wiki-p1.8 &  232k & 1.9\%& 2.4\%& 86.4\%& 41.0\%\\
  wiki-p6.0 &  737k& 2.8\%& 1.3\% & 82.2\%& 54.7\%\\
  wiki-mw-p1.6 & 589k& 4.7\% & 1.8\% & 83.8\%&59.4\%\\
 \bottomrule
 \end{tabular}}
 \caption{\label{table:CXA}Results on linking to lexical semantic resource: number of linked induced word senses, Coverage, ExtraCoverage, accuracy of our method and of the MFS baseline for our five datasets.}
\end{table}

\paragraph{\textbf{Results and discussion.}} In Table \ref{table:CXA} we present the results using the optimal parameter values (i.e. $th$=0.0 and $m$=5 of Algorithm \ref{alg:linking})\footnote{To optimize $m$, we prototyped our approach on a dev set consisting of a random sample of 300 senses, and studied the curves for  the number of linked induced senses to WordNet resp. BabelNet. The $th$ value was then selected as to maximize the accuracy.} For all datasets the number of linked senses, Coverage and ExtraCoverage are directly proportional to the number of entries in the dataset -- i.e., the finer the sense granularity, as given by a lower sense clustering $n$ parameter, the lower the number of mapped senses, Coverage and ExtraCoverage.

In general, we report rather low coverage figures: the coverage in WordNet is always lower than 50\% (30\% in one setting) and coverage in BabelNet is in all settings lower than 5\%. Low coverage is due to different levels of granularities between the source and target resource. Our target lexical resources, in fact, have very fine-grained sense inventories. For instance, BabelNet lists 17 senses of the word \term{python} including two (arguably obscure ones) referring to particular roller coasters. In contrast, word senses induced from text corpora tend to be coarse and corpus-specific. Consequently, the low coverage comes from the fact that we connect a coarse and a fine-grained sense inventory -- cf. also previous work \cite{FaralliNavigli13emnlp} showing comparable proportions between coverage and extra-coverage of automatically acquired knowledge (i.e., glosses) from corpora.

Finally, our results indicate differences between the order of magnitude of the Coverage and ExtraCoverage when linking to WordNet and BabelNet. This high difference is rooted in the cardinality of the two sense inventories: whereas BabelNet encompasses millions of senses, WordNet contains hundreds of thousands -- many of them not covered in our corpora.  Please note that an ExtraCoverage of about 3\% in BabelNet corresponds to about 300k novel senses. Overall, we take our results to be promising in that, despite the relative simplicity of our approach (i.e., almost parameter-free unsupervised linking), we are able to reach high accuracy figures in the range of around 87--95\% for WordNet and accuracies consistently above 80\% for BabelNet. This compares well against a random linking baseline that is able to achieve 44.2\% and 40.6\% accuracy on average when mapping to WordNet and BabelNet, respectively. Also, we consistently outperform the strong performance exhibited by the MFS baselines, which, in line with previous findings on similar tasks \cite{Suchaneketal:08,ponzetto09a} provide a hard-to-beat competitor. Thanks to our method, in fact, we are able to achieve an accuracy improvement over the MFS baseline ranging from 1.4\% to 14.3\% on WordNet mappings, and from 24.4\% to 45.4\% on BabelNet. Despite not being comparable, our accuracy figures are in the same ballpark as those reported by  \cite{navigli12} (cf.\ Table 1), who use a similar method for linking Wikipedia to WordNet.

\paragraph{\textbf{Error analysis.}} To gain insights into the performance of our approach, as well as its limitations, we performed a manual error analysis of the output on the WordNet mappings, identifying a variety of sources of errors that impact the quality of the output resource. These include:
\begin{itemize}[leftmargin=4mm]
\item \textbf{part-of-speech tagging errors}, which may produce wrong senses such as nonexistent `verbs' (e.g.\ \term{tortilla:VB}) (about 10\% of the errors);
\item \textbf{Hearst patterns errors} that may extract wrong hypernyms such as \term{issue} for the entry \term{emotionalism} (about 20\% of the errors); 
\item \textbf{linking errors} where the accuracy strongly depends on the granularity of senses and relationships of the target lexical resource (about 70\% of the errors).
\end{itemize}
More specifically, false positives are often caused by the selection of a synset that is slightly different from the most suitable one (i.e., semantic shift), whereas false negatives typically occur due to the lack of connectivity in the semantic network.

Even if the high values of the estimated accuracy (see Table \ref{table:CXA}) of our mapping approach indicate that we are generally performing well over all the classes of test examples (i.e., true positive, true negative, false positive and false negative), the performance figures exhibit a different order of magnitude between the count of true positives and true negatives.  True negatives are senses in the ExtraCoverage that we estimate to be correct new senses not contained in the reference lexical resource. For a sample of such senses we performed an additional manual analysis, and identified the following reasons that explain our generally high ExtraCoverage scores.

\begin{itemize}[leftmargin=4mm]
\item \textbf{Named entities and domain-specific senses} (about 40\% of the true negatives): true negative senses are due to correct new senses not contained in the target lexical resource. This holds in particular for WordNet, where encyclopedic content occurs in a spotty fashion in the form of a few examples for some classes;
\item \textbf{Sense granularity misalignment} (about 60\% of the true negatives): true negatives that derive from excessively fine clustering, and should have been combined with other senses to represent a more generic sense.
\end{itemize}

\subsection{Experiment 3: Typing of the Unmapped Induced Senses}  
\label{sec:exp3}

\paragraph{\textbf{Experimental setting.}} The high ExtraCoverage rates from Section \ref{sec:exp2} show that our resource contains a large number of senses that are not contained in existing lexical resources such as WordNet and BabelNet. Besides, high accuracy scores in the evaluation of the quality of the sense clusters from Section \ref{sec:exp1} seem to indicate that such extra items are, in fact, of high quality. Crucially for our purposes, information found among the extra coverage has enormous potential, e.g., to go beyond `Wikipedia-only' sense spaces. Consequently, we next evaluate our semantic typing component (Section \ref{subsec:typing}) to assess the quality of our method to include also these good extra clusters that, however, have no perfect mapping in the reference lexical resource (WordNet, BabelNet). 

Similarly to the experiments for the resource mapping (Section \ref{sec:exp2}), we manually create five test sets, one for each dataset  from Section \ref{sec:datasets}, by randomly selecting 300 unmapped PCZ items for each dataset, and manually identifying the most appropriate type of each induced sense amongst WordNet or BabelNet senses. Given these gold standards, performance is then computed as standard accuracy on each dataset.

\begin{table}[t]
\footnotesize
  \centering
 \begin{tabular}{lrcccc} 
 \toprule
  \multicolumn{6}{c}{\bf WordNet-linked}\\ 
  PCZ &  \#extra senses & w types & w/o types & Accuracy & MFS baseline \\ 
 \midrule
 news-p1.6 & 244k & 184k & 59k& 83.3\% & 80.7\%\\
 news-p2.3 & 316k & 226k & 90k& 91.4\% & 89.5\%\\
 wiki-p1.8 & 277k & 225k & 51k & 89.2\% & 89.0\%\\
 wiki-p6.0 & 1M & 675k & 487k & 81.2\% & 78.2\%\\
 wiki-p1.6-mwe & 683k & 538k & 144K & 78.8\% & 77.3\%\\
  \midrule
  \multicolumn{6}{c}{\bf BabelNet-linked}\\ 
  PCZ &  \#extra senses & w types & w/o types & Accuracy & MFS baseline \\ 
 \midrule
 news-p1.6 &  168k & 73k & 95k & 91.2\%& 87.2\%\\
 news-p2.3 &  225k & 89k & 135k & 90.3\%& 89.8\%\\
 wiki-p1.8 &  208k & 143k & 65k & 87.2\%& 85.0\%\\
 wiki-p6.0 &  1,4M & 278k & 1.1M & 41.2\%& 40.3\%\\
 wiki-p1.6-mwe &  552k & 342k  & 209k & 89.6\%& 88.0\%\\
 \bottomrule
 \end{tabular}
 \caption{\label{table:TS}Statistics and performance on typing unmapped PCZ items: number of induced senses counting for ExtraCoverage, number of typed and untyped induced senses, accuracy of our method and accuracy of the MFS baseline for our five datasets.}
\end{table}

\paragraph{\textbf{Results and discussion.}}  In Table \ref{table:TS} we report the statistics and the estimated accuracy for the task of typing the previously unmapped senses found among the ExtraCoverage. For each dataset and lexical resource, we report: the number of senses in the ExtraCoverage, the number of senses for which we inferred the type, the number of senses for which we were not able to compute a type, and the estimated accuracy for the types inferred by our method on the basis of either the links generated using our approach from Section \ref{subsec:linking}, or those created using the MFS linking baseline. The results show that accuracy decreases for those datasets with higher polysemy. In particular, we obtain a low accuracy of 41.2\% for the `wiki-p6.0' where the disambiguated thesaurus contains only a low number of related senses, resulting in sparsity issues. For the other settings, the accuracy ranges from 78.8\% to 91.4\% (WordNet) and from 87.2\% to 91.2\% (BabelNet). The MFS baseline accuracies of typing the un-mapped induced senses (see Section \ref{sec:exp2}) are lower, scoring 0.2\% to 2.7\% less accuracy for WordNet and 0.5\% to 4.2\% less accuracy for BabelNet: these results corroborate the previous ones on linking, where the MFS was shown to be a tough  baseline. Besides, the higher performance figures achieved by the MFS on typing when compared to linking indicate that the typing task has a lower degree of difficulty in the respect that popular (i.e., frequent) types provide generally good type recommendations.

\subsection{Experiment 4: Evaluation of Enriched Lexical Semantic Resources}  
\label{sec:semeval}

\paragraph{\textbf{Experimental setting.}} In our next experiment, we follow previous work \cite{navigli12} and benchmark the quality of our resources by making use of the evaluation framework provided by the Sem\-Eval-2007 task 16 \cite{cuadros-rigau:2007:SemEval-2007} on the ``Evaluation of wide-coverage knowledge resources''. This SemEval task is meant to provide an evaluation benchmark to assess wide-coverage lexical resources on the basis of a traditional lexical understanding task, namely Word Sense Disambiguation \cite{Navigli2009}. The evaluation framework consists of two main phases:
\begin{enumerate}[leftmargin=4mm]

\item \textbf{Generation of sense representations.} From each lexical resource, sense representations, also known as ``topic signatures'', are generated, which are sets of terms that are taken to be highly correlated with a set of target senses. In practice, a sense representation consists of a weighted vector, where each element corresponds to a term that is deemed to be related to the sense, and the corresponding weight quantifies its strength of association.

\item \textbf{WSD evaluation.} Next, sense representations are used as weighted bags of words in order to perform monolingual WSD using a Lesk-like method (cf. \cite{Lesk86}) applied to standard lexical sample datasets. Given a target word in context and the sense representations for each of the target word's senses, the WSD algorithm selects the sense with the highest lexical overlap (i.e., the largest number of words in common) between the sense representation and the target word's textual context.

\end{enumerate}
This SemEval benchmark utilizes performance on the WSD task as an indicator of the quality of the employed lexical resource. This approach makes it possible to extrinsically compare the quality of different knowledge resources, while making as few assumptions as possible over their specific properties -- this is because knowledge resources are simply viewed as sense representations, namely weighted bags of words. Besides, to keep the comparison fair, it uses a common and straightforward disambiguation strategy (i.e., Lesk-like word overlap) and a knowledge representation formalism (i.e., sense representations) that is equally shared across all lexical resources when evaluating them on the same reference dataset. Specifically, the evaluation is performed on two lexical sample datasets, from the Senseval-3~\cite{mihalcea2004senseval} and Sem\-Eval-2007 Task 17~\cite{pradhan2007semeval} evaluation campaigns. The first dataset has coarse-grained and fine-grained sense annotations, while the second contains only fine-grained annotations. In all experiments, we follow the original task formulation and quantify WSD performance using standard metrics of recall, precision and balanced F-measure.

\begin{table}[t]
\scriptsize
\begin{tabular}{lllll} 
\toprule
\bf PCZ ID & \bf WordNet ID & \bf PCZ Related Terms  & \bf PCZ Context Clues\\ 
\midrule

mouse:0 & mouse:wn1 & rat\text{:0}, rodent\text{:0}, monkey\text{:0}, ... &  rat:conj\_and, gray:amod, ...\\ 

mouse:1 & mouse:wn4 &  keyboard\text{:1}, computer\text{:0}, printer\text{:0} ... & click:-prep\_of, click:-nn, ....  \\ 

keyboard:0 & keyboard:wn1 &  piano\text{:1}, synthesizer\text{:2}, organ\text{:0} ... & play:-dobj, electric:amod, .. \\ 

keyboard:1 & keyboard:wn1 & keypad\text{:0}, mouse\text{:1}, screen\text{:1} ... & computer, qwerty:amod ... \\ 
\bottomrule
\end{tabular}

\caption{Sample entries of the hybrid aligned resource (HAR) for the words \term{mouse} and \term{keyboard}. Trailing numbers indicate sense identifiers. To enrich WordNet sense representations we rely on related terms and context clues.}
\label{tab:DDTex2}
\end{table}

Here, we use the SemEval task to benchmark the `added value' in knowledge applicable for WSD that can be achieved by enriching a standard resource like WordNet with disambiguated distributional information from our PCZs on the basis of our linking. To this end, we experiment with different ways of enriching WordNet-based sense representations with contextual information from our HAR. For each WordNet sense of a disambiguation target, we first build a `core' sense representation from the content and structure of WordNet, and then expand it with different kinds of information that can be collected from the PCZ sense that is linked to it (cf.\ Table \ref{tab:DDTex2}):
\begin{itemize}[leftmargin=4mm]

\item \textbf{WordNet.} The baseline model relies solely on the WordNet lexical resource. It builds a sense representation for each sense of interest by collecting synonyms and definition terms from the corresponding WordNet synset, as well as all synsets directly connected to it (we remove stop words and weigh words with term frequency).
    
\item \textbf{WordNet + Related (news).} We augment the WordNet-based representation with related terms from the PCZ. That is, if the WordNet sense is linked to a corresponding induced sense in our resource, we add all related terms found in the linked PCZ sense to the sense representation.
      
\item \textbf{WordNet + Related (news) + Context (news/wiki). } Sense representations of this model are built by taking the previously generated ones, and additionally including terms obtained from the context clues of either the news (+ Context (news)) or Wikipedia (+ Context (wiki)) corpora we use (see Section~\ref{sec:datasets}).

\end{itemize}
In the last class of models, we used up to 5\,000 most relevant context clues per word sense. This value was set experimentally: performance of the WSD system gradually increased with the number of context clues, reaching a plateau at the value of 5\,000. During aggregation, we excluded stop words and numbers from context clues. Besides, we transformed syntactic context clues to terms, stripping the dependency type, so they can be added to other lexical representations. For instance, the context clue \textsf{rat:conj\_and} of the entry \sense{mouse:0} was reduced to the feature \term{rat}.

\begin{table}[t]
\scriptsize
\begin{tabular}{ll} 
\toprule
\bf Model & \bf Sense Representation \\ \midrule
WordNet-only  &  \parbox{3.45in}{
 memory, device, floppy, disk, hard, disk, disk, computer, science, computing, diskette, fixed, disk, floppy, magnetic, disc, magnetic, disk, hard, disc,      storage, device} \\ \midrule
 
WordNet + Related (wiki) & \parbox{3.45in}{recorder, disk, floppy, console, diskette, handset, desktop, iPhone, iPod, HDTV, kit, RAM, Discs, Blu-ray, computer, GB, microchip, site, cartridge, printer, tv, VCR, Disc, player, LCD, software, component, camcorder, cellphone, card, monitor, display, burner, Web, stereo, internet, model, iTunes, turntable, chip, cable, camera, iphone, notebook, device, server, surface, wafer, page, drive, laptop, screen, pc, television, hardware, YouTube, dvr, DVD, product, folder, VCR, radio, phone, circuitry, partition, megabyte, peripheral, format, machine, tuner, website, merchandise, equipment, gb, discs, MP3, hard-drive, piece, video, storage device, memory device, microphone, hd, EP, content, soundtrack, webcam, system, blade, graphic, microprocessor, collection, document, programming, battery, keyboard, HD, handheld, CDs, reel, web, material, hard-disk, ep, chart, debut, configuration, recording, album, broadcast, download, fixed disk, planet, pda, microfilm, iPod, videotape, text, cylinder, cpu, canvas, label, sampler, workstation, electrode,  magnetic disc, catheter, magnetic disk, Video, mobile, cd, song, modem, mouse, tube, set, ipad, signal, substrate, vinyl, music, clip, pad, audio, compilation, memory, message, reissue, ram, CD, subsystem, hdd, touchscreen, electronics, demo, shell, sensor, file, shelf, processor, cassette, extra,       mainframe, motherboard, floppy disk, lp, tape, version, kilobyte, pacemaker, browser, Playstation, pager, module, cache, DVD, movie, Windows, cd-rom, e-book, valve, directory, harddrive, smartphone, audiotape, technology, hard disk, show, computing, computer science, Blu-Ray, blu-ray, HDD, HD-DVD, scanner, hard disc, gadget, booklet, copier, playback, TiVo, controller, filter, DVDs, gigabyte, paper, mp3, CPU, dvd-r, pipe, cd-r, playlist, slot, VHS, film, videocassette, interface, adapter, database, manual, book, channel, changer, storage} \\ \bottomrule
\end{tabular}
\caption{WordNet-only and PCZ-enriched sense representations for the fourth WordNet sense of the word \term{disk} (i.e., the `computer science' one): the `core' WordNet sense representation is additionally enriched with related words from our hybrid aligned resource.}
\label{tab:expansions}
\end{table}

Table~\ref{tab:expansions} shows a complete example from our dataset that demonstrate how, thanks to our HAR, we are able to expand WordNet-based sense representations with many relevant terms from the related terms of our sense-disambiguated PCZs. 

We compare our approach to four state-of-the-art systems: KnowNet \cite{cuadros2008knownet}, BabelNet, WN+XWN~\cite{cuadros-rigau:2007:SemEval-2007}, and NASARI. KnowNet builds sense representations based on snippets retrieved with a web search engine. We use the best configuration reported in the original paper (KnowNet-20), which extends each sense with 20 keywords. BabelNet in its core relies on a mapping of WordNet synsets and Wikipedia articles to obtain enriched sense representations: here, we consider both original variants used to generate sense representations, namely collecting all BabelNet synsets that can be reached from the initial synset at distance one (`BabelNet-1') or two (`BabelNet-2') and then outputting all their English lexicalizations. The WN+XWN system is the top-ranked unsupervised knowledge-based system of  Senseval-3 and Sem\-Eval-2007 datasets from the original competition~\cite{cuadros-rigau:2007:SemEval-2007}. It alleviates sparsity by combining WordNet with the eXtended WordNet~\cite{xwn2001}. The latter resource relies on parsing of WordNet glosses. For all these resources, we use the scores reported in the respective original publications. 

NASARI provides hybrid semantic vector representations for BabelNet synsets, which are a superset of WordNet. Consequently, we follow a procedure similar to the one we use to expand WordNet-only sense representations with information from our PCZs -- namely, for each WordNet-based sense representation we add all features from the lexical vector of NASARI that corresponds to it.\footnote{We used the version of lexical vectors (July 2016) featuring 4.4 million of BabelNet synsets, yet covering only 72\% of word senses of the two datasets used in our experiments.} 


Thus, we compare our method to three hybrid systems that induce sense representations on the basis of WordNet and texts (KnowNet, BabelNet, NASARI) and one purely knowledge-based system (WN+XWN). Note that we do not include the supervised TSSEM  system in this comparison, as in contrast to all other considered methods including ours, it relies on a large sense-labeled corpus.


\begin{table}[t]
 \centering
 \scriptsize
 \begin{tabular}{lrrrrrr} 
 \toprule
 \multicolumn{1}{}{} & \multicolumn{3}{l}{Senseval-3 fine-grained} & \multicolumn{3}{l}{SemEval-2007 fine-grained}  \\ 
  \multicolumn{7}{c}{}\\

Model & \multicolumn{1}{c}{Precision} & \multicolumn{1}{c}{Recall} & \multicolumn{1}{c}{F$_1$} & \multicolumn{1}{c}{Precision} & \multicolumn{1}{c}{Recall} &\multicolumn{1}{c}{F$_1$} \\ \midrule

Random&19.1&19.1&19.1&27.4&27.4&27.4\\
WordNet (WN)                 & 29.7 & 29.7 & 29.7 & 44.3 & 21.0 & 28.5 \\
WN + Related (news)      & \bf \underline{47.5} & \bf \underline{47.5} & \bf \underline{47.5} & 54.0&50.0&51.9 \\
WN + Related (news) + Context (news) & 47.2 & 47.2 & 47.2 & 54.8 & 51.2 & 52.9 \\
WN + Related (news) + Context (wiki) & 46.9 & 46.9 & 46.9 & \bf 55.2 & \bf 51.6 & \bf 53.4 \\ \midrule
BabelNet-1 & \bf 44.3 & \bf 44.3 & \bf 44.3& 52.2 & 46.3 & 49.1 \\
BabelNet-2 & 35.0 & 35.0 & 35.0& \bf \underline{56.9} & \bf \underline{53.1} & \bf \underline{54.9} \\
KnowNet & 44.1 & 44.1 & 44.1 & 49.5 & 46.1 & 47.7\\
NASARI (lexical vectors) & 32.3 & 32.2 & 32.2 & 49.3 & 45.8 & 47.5 \\
WN+XWN & 38.5 & 38.0 & 38.3 & 54.9 & 51.1 & 52.9 \\
 \bottomrule
 \end{tabular}
 \caption{\label{table:s3} Comparison of our approach to the state of the art unsupervised knowledge-based methods on the SemEval-2007 Task 16 (weighted setting). The best results per section (i.e., the ones using our resources vs.\ those from the previous literature) are boldfaced, the best results overall are underlined.}
 \label{tab:sota}
\end{table}

\paragraph{\textbf{Results and discussion.}} Table~\ref{tab:sota} presents results of the evaluation, which generally indicate the high quality of the sense representations in our hybrid aligned resource. Expanding WordNet-based sense representations with distributional information provides, in fact, a clear advantage over the original representation on both Senseval-3 and Sem\-Eval-2007 datasets. Using related words specific (via linking) to a given WordNet sense provides substantial improvements in the results. Further expansion of sense representations with context clues (cf. Table~\ref{tab:DDTex}) provides a modest improvement on the Sem\-Eval-2007 dataset only. Consequently, the results seem to indicate that context clues generally do not provide additional benefits over the expansions provided from the related terms of the linked PCZ items for task of generating sense representations.

On the Senseval-3 dataset, our hybrid models show better performance than all unsupervised knowledge-based approaches considered in our experiment. On the Sem\-Eval-2007 dataset instead, we perform on a par, yet slightly below BabelNet's best setting. Error analysis of the sense representations suggests that the extra performance of BabelNet on the SemEval data seems to derive from an aggressive graph-based expansion technique that leverages semantic relations harvested from Wikipedias in many languages -- cf.\ the overall lower performance obtained by collecting sense representations from all Babel synsets at depth 1 only (`BabelNet-1') vs.\ those that can be reached with two hops (`BabelNet-2'). This `joint multilingual' approach has also been shown to benefit WSD in general \cite{navigli12d}, and represents an additional source of semantic information not present in our resource (we leave multilinguality for future work).

The results generally indicate the high quality of our hybrid aligned resource in a downstream application scenario, where we show competitive results while being much less resource-intensive. This is because our method relies only on a relatively small lexical resource like WordNet and raw, unlabeled text, as opposed to huge lexical resources like BabelNet or KnowNet. That is, while our method shows competitive results better or on a par with other state-of-the-art systems, it does not require access to web search engines (KnowNet), the structure and content of a very large collaboratively generated resource and texts mapped to its sense inventory (BabelNet, NASARI), or even a machine translation system or multilingual interlinked Wikipedias (BabelNet). 

\paragraph{\textbf{Related work on unsupervised WSD using the induced sense inventory.}}

This article is focused on the hybrid aligned resource, e.g. a PCZ linked to a lexical resource, as in the \textit{knowledge-based} WSD experiment described above. However, the induced sense inventory is a valuable resource on its own and can be used to perform \textit{unsupervised knowledge-free} WSD. In this case, the induced sense representations, featuring context clues, related words and hypernyms, are used as features representing the induced senses. In our related experiments with sparse count-based~\cite{panchenko16b,panchenko2017unsup} and dense prediction-based~\cite{pelevina-EtAl:2016:RepL4NLP} distributional models we show that such unsupervised knowledge-free disambiguation models yield state-of-the-art results as compared to the unsupervised systems participated in SemEval 2013 and the AdaGram~\cite{bartunov2015breaking} sense embeddings model. An interactive demo that demonstrates our model developed in these experiments is described in Figure~\ref{fig:demo} and in~\cite{panchenko2017d}.

\section{Applications}  
\label{sec:applications}

Linked distributional disambiguated resources carry a great potential to positively impact many knowledge-rich scenarios. In this section, we leverage our resource for a few downstream applications of knowledge acquisition, namely: i) noise removal in automatically acquired knowledge graphs, and ii) domain taxonomy induction from scratch.

\subsection{Linking Knowledge Resources Helps Taxonomy Construction}
\label{sec:cm}

\paragraph{\textbf{Task definition.}} We examine a crucial task in learning a taxonomy (i.e., the \emph{isa} backbone of a lexical-semantic resource) from scratch \cite{bordea2015semeval,bordea2016semeval}, namely the induction of clean taxonomic structures from noisy hypernym graphs such as, for instance, those obtained from the extractions of hyponym-hypernym relations from text. In this task, we are given as input a list of subsumption relations between terms or, optionally, word senses -- e.g., those from our PCZs (Figure \ref{tab:DDTex}) -- which can be obtained, for instance, by exploiting lexical-syntactic paths \cite{Hearst1992,Snowetal:04}, distributional representations of words \cite{baroni12,roller14}  or a combination of both \cite{shwartz-goldberg-dagan:2016:P16-1}. Due to the automatic acquisition process, such lists typically contain noisy, inconsistent relations -- e.g., multiple inheritances and cycles -- which do not conform to the desired, clean hierarchical structure of a taxonomy. Therefore, the task of \emph{taxonomy construction} focuses on bringing order among these extractions, and on removing noise by organizing them into a directed acyclic graph (DAG) \cite{KozarevaetHovy:2010}.

\paragraph{\textbf{Related work.}}  State-of-the-art algorithms differ by the amount of human supervision required and their ability to respect some topological properties while pruning the noise. Approaches like those of \cite{KozarevaetHovy:2010}, \cite{velardi13} and \cite{Kapanipathietal:14}, for instance, apply different topological pruning strategies that require to specify the root and leaf concept nodes of the KB in advance -- i.e.,  a predefined set of abstract top-level concepts and lower terminological nodes, respectively. The approach of \cite{Farallietal:15} avoids the need of such  supervision with an iterative method that uses an efficient variant of topological sorting \cite{Tarjan:72} for cycle pruning. Such lack of supervision, however, comes at the cost of not being able to preserve the original connectivity between the top (abstract) and the bottom (instance) concepts. Random edge removal \cite{Farallietal:15}, in fact, can lead to disconnected components, a problem shared with the OntoLearn Reloaded approach \cite{velardi13}, which cannot ensure such property when used to approximate a solution on a large noisy graph.

\paragraph{\textbf{ContrastMedium algorithm.}} Links between heterogeneous knowledge resources, like those found within our hybrid aligned resource (Section \ref{sec:linking}), can be leveraged, together with a specialized algorithm, in order to advance the state of the art in taxonomy construction. To this end, we use ContrastMedium, a novel algorithm \cite{Farallietal:2017} that is able to extract a clean taxonomy from a noisy knowledge graph without needing to know in advance -- that is, having to manually specify -- the top-level and leaf concepts of the taxonomy, while preserving the overall connectivity of the graph. ContrastMedium achieves this by projecting the taxonomic structure from a \emph{reference taxonomy} (e.g., WordNet or the taxonomic \emph{isa} backbone of BabelNet) onto a \emph{target (noisy) hypernym graph} --  for instance, the graph built from the set of hypernym relations in our hybrid aligned resource (Section \ref{sec:workflow}) -- on the basis of links found between the two resources, e.g., those automatically generated using our method from Section \ref{sec:linking}.

\begin{figure}
\begin{floatrow}
\ffigbox{%
     \includegraphics[width=1.0\linewidth]{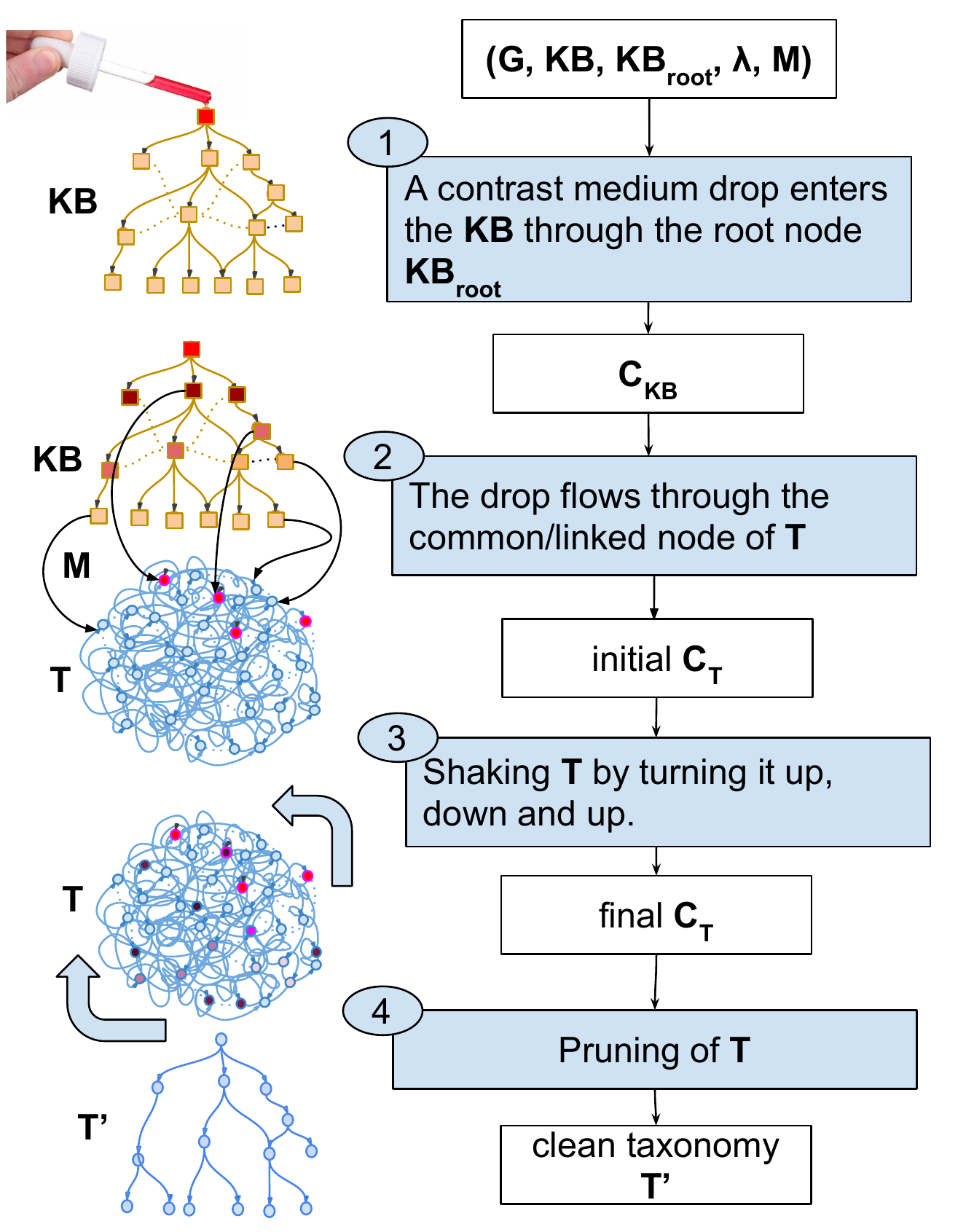}
}{%
  \caption{\label{fig:workflow-cm}The ContrastMedium (CM) algorithm for taxonomy construction.}%
}
\hspace{-9.5em}
\capbtabbox{%
\begin{tabular}{lrrr} 
\multicolumn{4}{l}{ContrastMedium}\\
& \multicolumn{1}{c}{$A_R$} &    \multicolumn{1}{c}{$A_M$} & \multicolumn{1}{c}{$A_L$}  \\
 news-p1.6 &  \textbf{98.9\%} & \textbf{98.3\%} & \textbf{99.3\%}\\ 
 news-p2.3 &  \textbf{98.7\%} & \textbf{98.7\%} & \textbf{99.9\%} \\ 
 wiki-p1.8 & \textbf{97.6\%} & \textbf{94.7\%} & \textbf{97.3\%}  \\ 
 wiki-p6.0 & \textbf{95.9\%} & \textbf{94.3\%} & \textbf{98.3\%}  \\ 
 &&&\\
 \cline{1-4}
  &&&\\
\multicolumn{4}{l}{Tarjan (baseline)}\\ 
 & \multicolumn{1}{c}{$A_R$} &    \multicolumn{1}{c}{$A_M$} & \multicolumn{1}{c}{$A_L$}  \\
 news-p1.6 &  93.3\% & 94.6\% & 95.3\%\\ 
 news-p2.3 &  95.7\% & 94.7\% & 95.6\%  \\ 
 wiki-p1.8 &  93.1\% & 87.3\% & 94.1\% \\ 
 wiki-p6.0 & 89.5\% & 90.1\% & 92.8\%  \\ 
 &&&\\
 &&&\\
 \end{tabular}
}{%
  \caption{\label{table:results1}Pruning accuracy of the CM.}%
}
\end{floatrow}
\end{figure}

Metaphorically, in the context of clinical analysis, a contrast medium (CM) is injected into the human body to highlight specific complex internal body structures (in general, the cardiovascular system). In a similar fashion, our approach, which is summarized in Figure \ref{fig:workflow-cm}, starts by detecting the topological structure of the reference taxonomy by propagating a certain amount of CM that we initially inject through its root node (step 1). The highlighted structure indicates the distance of a node with respect to the root, with the lowest values of CM indicating the leaf terminological nodes. The observed quantities are then transferred to corresponding nodes of the target hypernym graph by following the links between the two resources (step 2). Next, the medium is propagated by 'shaking' the noisy graph. We let the fluid reach all its components by alternating two phases of propagation: letting the CM flow via both incoming (`shake up') and outgoing (`shake down') edges (step 3). Finally, we use the partial order induced by the level of CM observed in each node to drive the pruning phase, and we `stretch' the linked noisy knowledge graph into a proper taxonomy, namely a DAG (step 4).

\paragraph{\textbf{Evaluation.}} We benchmark ContrastMedium by comparing the quality of its output taxonomies against those obtained with the state-of-the-art approach of \cite{Farallietal:15}. The latter relies on Tarjan's topological sorting, which iteratively searches for a cycle (until no cycle can be found) and randomly removes an edge from it. We applied the two approaches to our linked resources and evaluated the performance on a 3-way classification task to automatically detect the level of granularity of a concept. Pruning accuracy is estimated on the basis of a dataset of ground-truth judgments that were created using double annotation with adjudication from a random sample of 1,000 nodes for each noisy hypernym graph ($\kappa = 0.657$ \cite{fleiss1971measuring}). To produce a gold-standard, coders were asked to classify concepts from the random sample as: i) a root, top-level abstract concept -- i.e., any of \term{entity}, \term{object}, etc.\ and more in general nodes that correspond to abstract concepts that we can expect to be part of a core ontology such as, for instance, DOLCE \cite{DOLCE:2003}; ii) a leaf terminological node (i.e., instances such as \term{Pet Shop Boys}); or iii) a middle-level concept (e.g., \term{celebrity}), namely concepts not fitting into any of the previous classes. 

We compute standard accuracy for each of the three classes. That is, we compare the system outputs against the gold standards and obtain three accuracy measures: one for the root nodes ($A_R$), one for the nodes `in the middle' ($A_M$) and finally one for the leaf nodes ($A_L$).  In Table \ref{table:results1} we show some of the results of the evaluation. Thanks to ContrastMedium, we are able to achieve, even despite the baseline already reaching very high performance levels (well above 90\% accuracy), improvements of up to 6 percentage points, with an overall error reduction between around 40\% and 60\%. This performance improvements are due to the fact that ContrastMedium is able to: i) identify important topological clues among ground-truth taxonomic relations from the reference taxonomy, and ii) project them onto the noisy graph on the basis of the links found in the mapping between the two resources. That is, the availability of a mapping between knowledge resources helps us to project the supervision information from the clean source taxonomy into the target noisy graph without the need of further supervision. The reference taxonomy provides us with ground-truth taxonomic relations -- this renders our method as knowledge-based, not knowledge-free. However, the availability of resources like, for instance, WordNet for English or the multilingual BabelNet implies that these requirements are nowadays trivially satisfied. The mapping, in turn can be automatically generated with high precision using any of the existing solutions for KB mapping, e.g., our algorithm from Section \ref{sec:linking}, or by relying on ground-truth information from the Linguistic Linked Open Data cloud \cite{Chiarcos12}.

What is perhaps the most interesting bit in our approach is the fact that by combining our unsupervised framework for knowledge acquisition from text (Section \ref{sec:workflow}) with ContrastMedium we are able to provide an \emph{end-to-end solution for high-quality, unsupervised taxonomy induction from scratch}, i.e., without any human effort.

\subsection{Inducing Taxonomies from Scratch using the Hybrid Aligned Resource}

\paragraph{\textbf{Task definition.}} We now look at how ContrastMedium can be used as component within a larger system to enable end-to-end taxonomy acquisition from text. In general, taxonomy learning from scratch \cite{bordea2015semeval,bordea2016semeval} consists of the task of inducing a hypernym hierarchy from text alone \cite{Biemann05}: this typically starts with an initial step of finding hypernymy relations from texts, which is followed by a taxonomy construction phase in which local semantic relations are arranged together within a proper global taxonomic structure (cf.\  previous Section \ref{sec:cm}).

\paragraph{\textbf{Related work.}}  Existing approaches like \cite{KozarevaetHovy:2010} (among others) use Hearst-like patterns~\cite{Hearst1992} to bootstrap the extraction of terminological sister terms and hypernyms. Instead, in~\cite{velardi13} the extraction of hypernymy relations is performed with a classifier, which is trained on a set of manually annotated definitions from Wikipedia~\cite{Navigli:2010:LWL:1858681.1858815}, being able to detect definitional sentences and to extract the definiendum and the hypernym. In these systems, the harvested hypernymy relations are then arranged into a taxonomy structure, e.g., using cycle pruning and `longest path' heuristics to induce a DAG structure~\cite{KozarevaetHovy:2010} or by relying on  a variant of Chu-Liu Edmonds' optimal branching algorithm~\cite{velardi13}. In general, all such lexical-based approaches suffer from the limitation of not being sense-aware, which results in spurious taxonomic structures. Now, our fully disambiguated sense inventories could potentially overcome this problem and enable to a step forward towards the induction of high-quality, full-fledged taxonomies. In fact, we now show how the linked/semantic nature of our resources enables the development of a complete approach for taxonomy induction from scratch that achieves state-of-the-art performance with virtually no explicit supervision.

\paragraph{\textbf{Using hybrid aligned resource to learn taxonomies from scratch.}}  

\begin{figure}
     \includegraphics[width=1.0\linewidth]{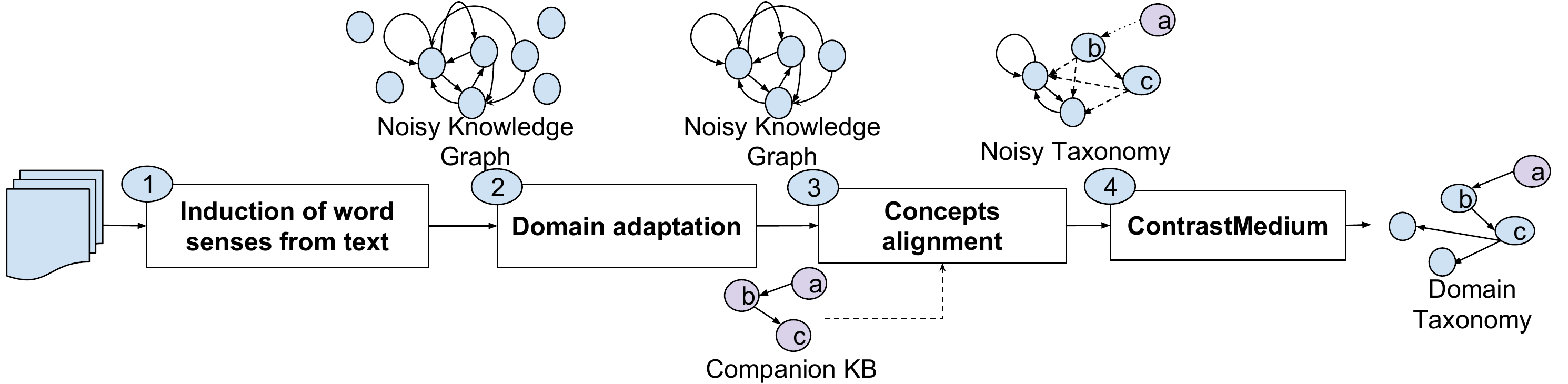}
     \caption{Our full end-to-end pipeline for taxonomy induction from scratch.}\label{fig:inductionpipe2}
\end{figure}

We now focus on the task of taxonomy induction by exploiting our hybrid aligned resource. Our approach is based on a five-stages pipeline (see Figure \ref{fig:inductionpipe2}):
\begin{enumerate}[leftmargin=4mm]
 \item create proto-conceptualizations (PCZs) as described in Section \ref{sec:DDTdesc};
 \item filter out-of-domain concepts from our PCZs on the basis of domain-terminology-based heuristics. We construct domain-specific PCZs for a target domain by a simple lexical filtering. First, we build an extended lexicon of each domain on the basis of a seed vocabulary of the domain -- i.e., domain terminologies such as those provided from the TExEval challenge (see below). Namely, for each seed term, we retrieve all semantically similar terms on the basis of the PCZ;
  \item build a hybrid aligned resource by linking the PCZs to a companion taxonomy (e.g., WordNet, BabelNet, etc.) based on the methods from Section \ref{sec:linking}; 
 \item build a noisy hypernym graph by taking the union of the hypernym relations found within our PCZs;
 \item apply ContrastMedium (Section \ref{sec:cm}) to remove noise from the graph and obtain a proper taxonomic structure.
\end{enumerate}
That is, the combination of all our methods we presented so far provides us with a full end-to-end pipeline for taxonomy induction from scratch. Arguably, our approach is unsupervised in that it does not require any explicit human effort other than the knowledge encoded within the reference lexical resources. Both PCZs and links to reference knowledge resources are automatically induced in an unsupervised way. Moreover, links to existing lexical resources, as used in ContrastMedium, provide us with a source of knowledge-based supervision that is leveraged to clean PCZs and turn them into full-fledged taxonomies. More precisely, our framework is fully unsupervised up to the linking part. However, unsupervised linking to a knowledge base and using the knowledge base for taxonomy construction indeed requires the knowledge base itself. To this end, we use freely available resources like WordNet and BabelNet. Given the linking, we can exploit the knowledge from these lexical resources, together with a knowledge-based method (ContrastMedium), without the need for additional human effort or supervision. That is, the fact that these lexical knowledge resources already exist and are publicly available implies that we can apply our framework with no extra human intervention.

\paragraph{\textbf{Experiments.}} We use the evaluation benchmark from the most recent edition of the TExEval challenge (SemEval 2016 - task 13)~\cite{bordea2016semeval}. Our experimental setting consists of the following components:
\begin{itemize}[leftmargin=4mm]
 
 \item \textbf{Three gold-standard taxonomies}, namely the \textsc{FOOD}'s sub hierarchy of the Google products taxonomy\footnote{\url{http://www.google.com/basepages/producttype/taxonomy.en-US.txt}}, as well as the  NASEM\footnote{\url{http://sites.nationalacademies.org/PGA/Resdoc/PGA_0445}} and EuroVoc\footnote{\url{http://eurovoc.europa.eu/drupal/}} taxonomies of \textsc{Science}.
  
 \item \textbf{The task baseline}, which induces the taxonomy structure only from relations between compound terms such as \term{juice}, \term{apple juice} by applying simple substring inclusion heuristics. This baseline approach does not leverage any external or background knowledge and only uses the input domain terminology.

\item \textbf{The Cumulative Fowlkes\&Mallows evaluation measure} (CF\&M): this enables the comparison of a system taxonomy against a gold standard at different levels of depth of the taxonomy, as obtained by penalizing errors at the highest cuts of the hierarchy \cite{VelardiNFR12}.

 \item \textbf{The task participant's systems}: (1) the JUNLP system~\cite{DBLP:conf/semeval/MaitraD16} makes use of two string inclusion heuristics combined with information from BabelNet; (2) the NUIG-UNLP system~\cite{DBLP:conf/semeval/Pocostales16} implements a semi-supervised method that finds hypernym candidates by representing them as distributional vectors~\cite{mikolov2013linguistic}; (3) the QASSIT system~\cite{DBLP:conf/semeval/CleuziouM16} is a semi-supervised methodology for the acquisition of lexical taxonomies based on genetic algorithms. It is based on the theory of pretopology~\cite{Aluja2012} that offers a powerful formalism to model semantic relations; (4) our task-winning TAXI system~\cite{panchenko2016taxi} that relies on combining two sources of evidence: substring matching and Hearst-like patterns. Hypernymy relations are extracted from Wikipedia, GigaWord, ukWaC, a news corpus and the CommonCrawl, as well as from a set of focused crawls; (5) the USAAR system~\cite{DBLP:conf/semeval/TanBG16} exploits the hypernym endo/exocentricity~\cite{nla.cat-vn1501373} as a practical property for hypernym identification.
 
 \item \textbf{A reference taxonomy}: we use WordNet for evaluation purposes by treating it the same way as any other participant system's output.

\end{itemize}
\begin{table}
\footnotesize
\centering
\begin{tabular}{p{7cm}rrr}
\hline
& \multicolumn{1}{c}{Google} & \multicolumn{1}{c}{NASEM} & \multicolumn{1}{c}{EuroVoc} \\
\hline
\multicolumn{1}{l}{System} & \multicolumn{1}{c}{\textsc{Foods}} & \multicolumn{1}{c}{\textsc{Sciences}} & \multicolumn{1}{c}{\textsc{Sciences}} \\
\hline
Baseline & 0.0019 &  0.0163& 0.0056 \\
JUNLP                             & 0.2608 &  0.1774 &  0.1373 \\
NUIG-UNLP                         & --  &  0.0090 &  0.1517 \\     
QASSIT                            & --   &  0.5757 &  0.3893  \\
TAXI                              & 0.2021 &  0.3634 &  0.3893  \\
USAAR                             & 0.0000 &  0.0020 &  0.0023  \\
WordNet                 & 0.5870 &  0.5760 &  0.6243  \\
\hline
Our approach              & \textbf{0.6862} &  \textbf{0.7000} & \textbf{0.8157}  \\
\hline
\end{tabular}
\caption{Comparison based on the SemEval 2016 task 13 benchmark for \textsc{Foods} and \textsc{Sciences} domains. We report the Cumulative Fowlkes\&Mallows measure.\label{table:Semeval}}
\end{table}
In Table~\ref{table:Semeval} we report the results on the SemEval gold standards. Our approach significantly ($\chi^2$ test, $p<.01$) outperforms all the other systems in all domains (i.e., Google \textsc{Food}, NASEM \textsc{Science} and EuroVoc \textsc{Science}), as well as the ground-truth taxonomy provided by WordNet. More importantly, the results indicate the overall robustness of our approach: that is, leveraging distributional semantics and symbolic knowledge (i.e., through linking to reference lexical resources) together is able to outperform not only the WordNet gold standard, which has limited coverage for fine-grained specific domains like these, but also the SemEval task participants, which all rely in some way or another on simple, yet powerful substring heuristics.

\section{Conclusions}
\label{sec:conclusions}

We have presented a framework for enriching lexical semantic resources, such as WordNet, with distributional information. Lexical semantic resources provide a well-defined semantic representation, but typically contain no corpus-based statistical information and are static in nature. Distributional semantic methods are well suited to address both of these problems, since models are induced from (in-domain) text  and can be used to automatize the process of populating ontologies with new concepts. However, distributional semantic representations based on dense vectors have also major limitations in that they are uninterpretable on the symbolic level. By linking these representations to a reference lexical resource, we can interpret them in an explicit way by means of the underlying relational knowledge model.

We provided a substantial investigation on enrichment of lexical resources with distributional semantics and evaluated the results in  intrinsic and extrinsic ways showing that the resulting hybrid sense representations can be successfully applied to a variety of tasks that involve lexical-semantic knowledge. We tested the quality of the hybrid resources generated by our framework with a battery of intrinsic evaluations. Additionally, we benchmarked the quality of our resource in a knowledge-based word sense disambiguation setting, where we showed that our arguably low-resource approach -- in that we rely only on a small lexical resource like WordNet and raw, unlabeled text -- reaches comparable quality with BabelNet, which in contrast is built on top of large amounts of high-quality collaborative content from Wikipedia. Finally, by combining distributional semantic vectors with links to a reference lexical resources, we are able to pave the way to the development of new algorithms to tackle hard, high-end tasks in knowledge acquisition like taxonomy cleaning and unsupervised, end-to-end taxonomy learning.
 
We believe that the hybrid lexical resources developed in our work will benefit high-end applications, e.g.\ ranging from entity-centric search \cite{lin12,schuhmacher15a} all the way through full-fledged document understanding \cite{Rospocher2016}. 

\paragraph{\textbf{Downloads}.} We release all resources produced in this work under CC-BY 4.0 License\footnote{\url{https://creativecommons.org/licenses/by/4.0/}}: i) the PCZs resulting from our first experiment (Section \ref{sec:exp1}); ii) following the guidelines in \cite{Mccraeetal2014}, we created an RDF representation to share the mapping between our PCZs and lexical knowledge graphs (i.e., WordNet and BabelNet) (see Section  \ref{sec:exp2}) in the Linked Open Data Cloud; iii) the types of the unmapped PCZ senses produced in the third experiment (see Section \ref{sec:exp3}). All datasets, evaluation judgments, source code, and the demo can be accessed via \url{http://web.informatik.uni-mannheim.de/joint}.

\nocite{riedl2016phd}
\bibliographystyle{fullname2}
\bibliography{prontobib}


\end{document}